\documentclass[a4paper,fleqn]{cas-sc}

\usepackage[numbers]{natbib}
\usepackage{xcolor}
\usepackage{soul}
\usepackage{lineno}
\usepackage[caption=false]{subfig}
\usepackage{tabularx}
\def\tsc#1{\csdef{#1}{\textsc{\lowercase{#1}}\xspace}}
\tsc{WGM}
\tsc{QE}

\begin{document}
\let\WriteBookmarks\relax
\def\floatpagepagefraction{1}
\def\textpagefraction{.001}

\shorttitle{Residual Attention for Vision Transformers}    

\shortauthors{Anxhelo Diko, Danilo Avola, Marco Cascio, Luigi Cinque}  

\title [mode = title]{ReViT: Enhancing Vision Transformers Feature Diversity with Attention Residual Connections}



%

\author[1]{Anxhelo Diko}
\ead{diko@di.uniroma1.it}


\credit{Conceptualization, Methodology, Software, Writing}

\affiliation[1]{organization={Department of Computer Science, Sapienza University of Rome},
            addressline={Via Salaria 113}, 
            city={Rome},
            postcode={00198}, 
            state={Italy},
            country={Italy}}
\affiliation[2]{organization={Department of Law and Economics, University of Rome UnitelmaSapienza},
            addressline={Piazza Sassari 4}, 
            city={Rome},
            postcode={00161}, 
            state={Italy},
            country={Italy}}

\author[1]{Danilo Avola}
\cormark[*]
\cortext[cor1]{Corresponding Author}
\corref{cor1}
\ead{avola@di.uniroma1.it}
\credit{Supervision, Conceptualization, Writing}

\author[1,2]{Marco Cascio}
\ead{cascio@di.uniroma1.it}
\credit{Conceptualization, Methodology, Writing}

\author[1]{Luigi Cinque}
\ead{cinque@di.uniroma1.it}
\credit{Supervision}

\begin{abstract}
Vision Transformer (ViT) self-attention mechanism is characterized by feature collapse in deeper layers, resulting in the vanishing of low-level visual features. However, such features can be helpful to accurately represent and identify elements within an image and increase the accuracy and robustness of vision-based recognition systems. Following this rationale, we propose a novel residual attention learning method for improving ViT-based architectures, increasing their visual feature diversity and model robustness. In this way, the proposed network can capture and preserve significant low-level features, providing more details about the elements within the scene being analyzed. The effectiveness and robustness of the presented method are evaluated on five image classification benchmarks, including ImageNet1k, CIFAR10, CIFAR100, Oxford Flowers-102, and Oxford-IIIT Pet, achieving improved performances. Additionally, experiments on the COCO2017 dataset show that the devised approach discovers and incorporates semantic and spatial relationships for object detection and instance segmentation when implemented into spatial-aware transformer models.
\end{abstract}


\maketitle

\section{Introduction}\label{s:intro}

Nowadays, automatic visual recognition systems have become increasingly popular as powerful support tools for a wide range of vision-related applications, e.g., object detection and tracking \cite{chen2022@swipnet, gao2023@tracking}, image analysis and classification \cite{zheng2017@classification}, or scene segmentation \cite{yin2021@agunet} and understanding \cite{gonzales2016@visual}. Such advancements were initially fueled by the prowess of Convolutional Neural Networks (CNNs) in extracting multi-scale localized features \cite{li2021survey,laith2021cnn}. However, CNNs are bounded by limited receptive fields, hindering their ability to model arbitrary relationships among distant image regions. To address these limitations, several works started exploring vision transformer (ViT) architectures \cite{dosovitskiy2021ViT,hugo@aug} built upon the self-attention mechanisms, which unlocks the modeling of extensive dependencies between tokens within a data sequence \cite{ashish@2017attention}. Adapting with such a mechanism, in contrast to CNNs, engineered for visual inputs with built-in inductive biases such as locality and translation equivariance \cite{d2021convit}, ViT treats images as a sequence of local patches. However, despite using such an unconventional approach, ViT has propelled computer vision to new heights, showcasing the potential of a unified architecture for diverse tasks and modalities \cite{girdhar2023imagebind}.

\begin{figure}
\centering
\includegraphics[width=0.6\textwidth]{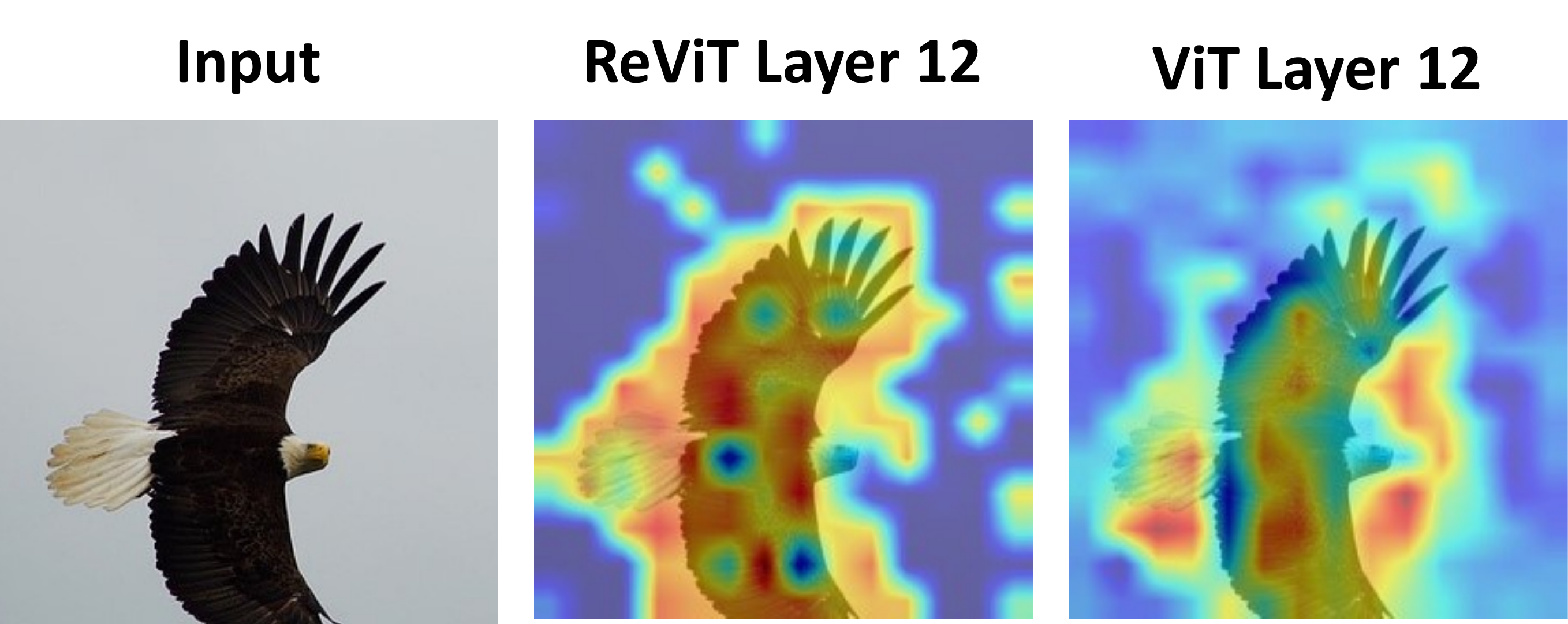}
 \caption{Illustration of feature maps learned from ReViT and ViT obtained using GradCAM \cite{ramprasaath2017@gcam} algorithm.}
 \label{fig:into}
\end{figure}

 While ViTs have revolutionized computer vision, their reliance on vast training datasets (hundreds of millions to billions of samples \cite{dosovitskiy2021ViT}) limits their applicability to smaller datasets impeding their development for domain-specific applications \cite{li2022locality}. Such reliance has effectively rendered datasets like ImageNet1K \cite{olga2015ImageNet}, formerly considered a rigorous benchmark for training vision models, a legacy standard primarily used for assessing classification performance. However, even though modern computing power allows for large-scale training, the ability to converge on smaller datasets remains crucial. This is especially important for enabling real-world applications dealing with specialized domains like medical imaging \cite{zhu2021hard}, aerial imaging \cite{kulkarni2023aerial}, etc., where data collection is expensive and limited. Moreover, data in these domains often diverges significantly from the data used to pre-train large transformer models, hindering their effectiveness \cite{li2022locality}. These limitations introduce important challenges to ViT architectures, emphasizing the need for modern solutions that can also unlock its ability to converge in smaller-scale datasets.

 According to the literature, the primary rationale for such limitation is twofold: 1) the absence of inherent inductive bias \cite{d2021convit,dosovitskiy2021ViT,Liu@2021swin}, and 2) feature collapsing \cite{tang2021augmented}. The former arises from the self-attention mechanism, which is explicitly designed for images and, therefore, lacks inherent assumptions about the spatial structure of the data. However, despite hindering the performance of transformers on restricted data, this can also be considered a treat since it is crucial in enabling the flexibility to generalize across modalities. On the other hand, the second and most critical rationale, feature collapsing, refers to the degradation of information within transformer architectures as features progress through the transformer network. Specifically, transformers are typically composed of stacked computational blocks, including multi-head self-attention layers (MHSA), multi-layer perceptron (MLP), layer normalization operators (LN), and residual connections. As the input features flow forward through these blocks, they become indistinguishable due to inter-token feature aggregation within the self-attention mechanism \cite{tang2021augmented}. This causes feature collapsing, resulting in overly-global representations that encode only global information, omitting crucial local details. Such a phenomenon hinders the network's ability to extract robust representations necessary for vision tasks like object classification or detection \cite{hugo@aug,tang2021augmented}.

In their attempts to improve the vision transformer architecture, many prior works have proposed interesting solutions by implicitly or explicitly targeting the above limitations. For instance, in \cite{hugo@aug}, authors propose using heavy data augmentation and knowledge distillation from CNNs to improve learning efficiency. Following, several other works try to inspire from the effectiveness of CNNs and morph the transformer architecture by inducing CNN-like initialization bias \cite{d2021convit}, windowed processing \cite{Liu@2021swin}, and multi-scale hierarchical transformer architectures \cite{Li@2022MViT,Liu@2021swin} able to mimic the convolution operation using the attention mechanism. However, while these approaches might be effective, they diverge from the natural ability of transformers to generalize across modalities and enable the unification of software and hardware design. In such regards, in \cite{tang2021augmented}, authors propose to improve the efficiency of transformers targeting features collapsing through augmented residual connections between features of consecutive attention blocks. This is the first work addressing the core issue of transformers related to learning efficiency. Nevertheless, the proposed approach introduces a non-negligible computational burden as it requires the addition of several MLP layers parallel to each transformer block, making it less suitable for real-world applications. Moreover, such an attempt fails to address the feature collapsing at its core: self-attention over-globalization.

In pursuit of addressing these transformer limitations and the shortcomings of previous approaches, this work presents the \textbf{Re}sidual attention \textbf{Vi}sion \textbf{T}ransformer (ReViT)\footnote{\href{https://github.com/ADiko1997/ReViT}{https://github.com/ADiko1997/ReViT}} architecture. Inspired by the human brain's hierarchical processing of visual information, where simple cells aggregate electrical impulses to transmit signals to specialized cells \cite{huff2021neuroanatomy}, ReViT tackles the challenge of feature collapsing observed in conventional ViTs. It does so by introducing a novel attention residual learning technique contrasting the over-globalization of self-attention, a phenomenon that fuels feature collapsing as highlighted in \cite{tang2021augmented,dosovitskiy2021ViT,d2021convit}.
Specifically, given standard vision transformer blocks, attention information is passed between consecutive layers and integrated with the attention information of the current layer in the form of a residual connection using a permutation invariant aggregation function. This connection facilitates the transmission and accumulation of attention-related information from shallower to deeper layers, enabling the ability to identify low-level features while preserving the original ability of transformers to extract global context. Consequently, by empowering the attention mechanism with the ability to capture such features, the resulting learned representations are expected to exhibit greater feature diversity.  Additionally, due to its simple nature, ReViT does not morph the core transformer architecture preserving its innate capabilities and computational load. 

In order to assess the efficacy of incorporating residual attention into ViT (i.e., ReViT), we conducted a comprehensive empirical analysis. Our evaluation spanned five standard image classification benchmarks, namely ImageNet1K \cite{olga2015ImageNet}, CIFAR-10 \cite{krizhevsky2009learning}, CIFAR-100 \cite{krizhevsky2009learning}, Oxford Flowers-102 \cite{Nilsback2008Flower}, and Oxford-IIIT Pet \cite{parkhi2012cats}. These datasets represent a range of scales and complexities, including medium-to-small scale datasets that better reflect real-world scenarios with limited data availability. Our findings demonstrate consistent performance improvements with ReViT compared to the standard ViT across all benchmarks, even under challenging conditions of image alterations that test the model's robustness.
Moreover, quantitative analysis of the attention maps, using the non-locality metric \cite{d2021convit}, reveals that ReViT effectively preserves more local information than standard ViTs, showcasing its effectiveness in preserving feature diversity. Further qualitative analysis visually illustrates ReViT's superior ability to integrate low-level features into its learned representations, as depicted in Fig. \ref{fig:into}. Finally, we established that the proposed residual learning strategy can seamlessly integrate into multiscale transformer architectures like the Multiscale Vision Transformer v2 (MViTv2) \cite{Li@2022MViT} and the Shifted-window transformer (Swin) \cite{Liu@2021swin}, verifying its effectiveness in various tasks, including image classification on ImageNet1K, object detection, and instance segmentation on COCO2017 \cite{lin2014coco}, obtaining improved performances.

In summary, the major contributions of this paper are:
\begin{enumerate}
    \item This work presents ReViT, a novel Vision Transformer architecture designed to tackle the limitation of feature collapsing. ReViT leverages the concept of residual attention, which, to the best of our knowledge, has not been previously explored.
    \item Demonstrate empirically and theoretically that residual attention effectively contrasts feature collapsing on ViT architectures.
    \item Extensive experimental results in several benchmarks assess the residual attention mechanism in enhancing performance across visual tasks (image classification, semantic segmentation, and object detection) on single and multi-scale architectures, demonstrating its effectiveness and versatility.
\end{enumerate}

The rest of this paper is organized as follows. Section 2 introduces relevant work that inspired this study. Section 3 comprehensively describes important background information and the proposed method. Section 4 describes the experiments performed to validate the proposed approach quantitatively and qualitatively. Section 5 reports an ablation study on the residual attention.  Finally, Section 6 draws the final conclusions from this study.

%
%
%
%

\section{Related Work}\label{s:relwork}

\subsection{Vision Transformers}\label{s:relwork}
In visual recognition tasks, standard CNN-based models are currently the most powerful algorithms for image analysis \cite{yuanduo@2022resDnet, Wang2017res}; even so, these networks are designed to capture short-range dependencies using small convolution filters that only look at small image patches at a time. However, due to its self-attention mechanisms, the transformer architecture was recently applied to process images, capturing long-range dependencies between image patches, thus enabling better modeling of complex visual relationships. Therefore, the ViT \cite{dosovitskiy2021ViT} neural model was introduced, splitting the image into a grid of local patches fed through a stack of transformer blocks, which use self-attention to encode the global relationships between the different patches. To enhance ViT capabilities, Touvron et al. \cite{hugo@aug} proposed the Data-efficient Image Transformers (DeIT) using a knowledge distillation technique that transfers the knowledge learned by a larger CNN teacher model to a transformer model during training. This helps to improve ViT's generalization performance. However, despite the promising results, extracting low-level image features becomes a significant challenge. To address such an issue, Han et al. \cite{han2021TnT} introduced a new kind of neural architecture that divides image patches into sub-patches and later employs an extra transformer block to analyze their relationship. For the same reason, Liu et al. \cite{Liu@2021swin} presented the Swin transformer, a multiscale architecture operating in local windows with a shifted window policy to extract spatial cross-window relationships between patches. Trying to address the shortcomings of local-window self-attention in vision transformers, such as limited receptive field and weak modeling capabilities, Qiang et al. \cite{qiang2022mixformer} introduced the MixFormer model integrating local-window self-attention with depth-wise convolution in a parallel configuration, enhancing the model's ability to capture both intra-window and cross-window connections. Moreover, in literature, several works proposed enhancing local feature representations by aggregation strategy \cite{yuan2021T2T} and integrating local and global attention layers \cite{chu2021twins, lin2022cat}. Aiming to extract features at different levels of granularity and resolution, Li et al. \cite{Li@2022MViT} proposed the MViTv2 to capture local and global image features using a hierarchical architecture with multiple scales, which can improve the model's ability to define and recognize objects within a scene in different sizes and orientations. Differently, to improve the performance of ViT-based models for large-scale image classification, Wang et al. \cite{wang2022kvt} introduced the kNN search inside the attention mechanism to reduce the impact of noisy or irrelevant features by exploiting the locality of patches to extract meaningful information, thus resulting in overall better feature representation. Again, Yu et al. \cite{Yu_2022_BMVC@BOAT} proposed the Bilateral lOcal Attention vision Transformer (BOAT) based on an attention mechanism that integrates the common image-space local attention with feature-space local attention. The latter can compute attention among relevant image patches even if they are not close to each other within the image plane. Therefore, this type of attention is a natural compensation to image-space local attention, which may miss capturing meaningful relationships between existing patches across different local windows. However, kNN-based and BOAT models require significant computational resources, limiting their practical applicability. More recently, still recognizing the limitations of existing methods that focus primarily on local or regional feature extractions, Ting et al. \cite{ting2023dualvit} proposed the Dual Vision Transformer (Dual-ViT) that integrates two distinct pathways: a semantic pathway for global feature compression and a pixel pathway for local detail enhancement. This design strategy optimizes the self-attention mechanism by balancing the computation between global semantic understanding and local feature refinement, reducing the computational load while enhancing the model performance. Investigating multi-scale feature modeling, Nie et al. \cite{nie2024scopevit} proposed ScopeViT, an efficient scale-aware vision transformer model that enhances the ViT capacity to perceive and integrate visual tokens at different scales by embedding them with varying receptive fields into distinct attention heads. This model is characterized by two lightweight attention mechanisms facilitating varied scale perception across the network and optimizing the computational cost by reducing the token count in attention processes. Again, Tang et al. \cite{tang2024CATNet} proposed the CATNet network employing the Multi-dimensional Convolutional Attention (MCA) module, enhancing the model's capability to process multi-scale features, and the Dual Attention Transformer (DAT) module, improving the accuracy of pixel-level predictions. Also, this network architecture tries to maintain high accuracy while significantly reducing the computational load by integrating convolutional attention mechanisms with transformer-based approaches. This work differs from the previous ones in not seeking to make transformers resemble more CNNs but proposing a novel residual connection that helps improve feature diversity on ViTs and improve their representation capabilities.



\subsection{Residual Connections and Feature Diversity}\label{ss:feat_div}
Traditional downstream ViT models are characterized by deep architectures, which typically result in reduced feature diversity as the depth increases. This limitation hinders their ability to represent information effectively. Residual connections, pioneered by He et al. \cite{he2016ResNet} presenting the ResNet architecture, are a well-established method to alleviate the vanishing gradient problem in deep networks and maintain feature diversity by allowing the unimpeded flow of gradients through the network. These connections create a direct pathway for earlier layers' features to impact deeper layers, enhancing the model's ability to learn diverse feature representations as the network goes deeper for downstream prediction. Inspired by these effects, residual connections between consecutive blocks have also been incorporated into the transformer architecture \cite{ashish@2017attention}, and its vision homologue ViT \cite{dosovitskiy2021ViT}.
Motivated by its impact, Tang et al. \cite{tang2021augmented} proposed an augmented residual connection scheme in ViT models that enhances the flow of information between layers through multiple parallel alternative paths. These shortcuts, essentially parametrized linear projection sequences, bypass the attention mechanism within the transformer block. This augmentation boosts feature diversity on the deeper layers of ViT. In contrast, while variants of ViT that bear similarities to CNNs, such as MViTv2 or Swin, are less affected by the reduction in feature diversity, it is essential to note that these models are primarily designed for image-specific tasks, deviating from the original ViT goal of achieving generalization across a wide range of tasks and modalities.
Furthermore, d'Ascoli et al. \cite{d2021convit} introduced a novel attention layer called the gated positional attention layer in ViTs. Equipped with a soft inductive bias to mimic the behavior of CNNs, this layer improves the feature diversity of ViT models. However, while these methods contribute to increased feature diversity, they either require adding significant computational load or alterations to the standard ViT workflow by changing the composition of the transformer block. 
For these reasons, the proposed method offers an agile solution. It increases feature diversity in ViT architectures by forwarding attention among transformer layers with minimal additional parameters and seamlessly integrating into the original ViT workflow. Additionally, contrary to \cite{tang2021augmented}, the proposed method addresses the issue of feature diversity by contrasting attention over globalization, which is the main catalyst in feature collapsing rather than propagating past features.

\section{Background and Methodology} \label{s:feat_coll_loc}
\subsection{Vision Transformer Layer}
\begin{figure}
\centering
\includegraphics[width=0.5\textwidth]{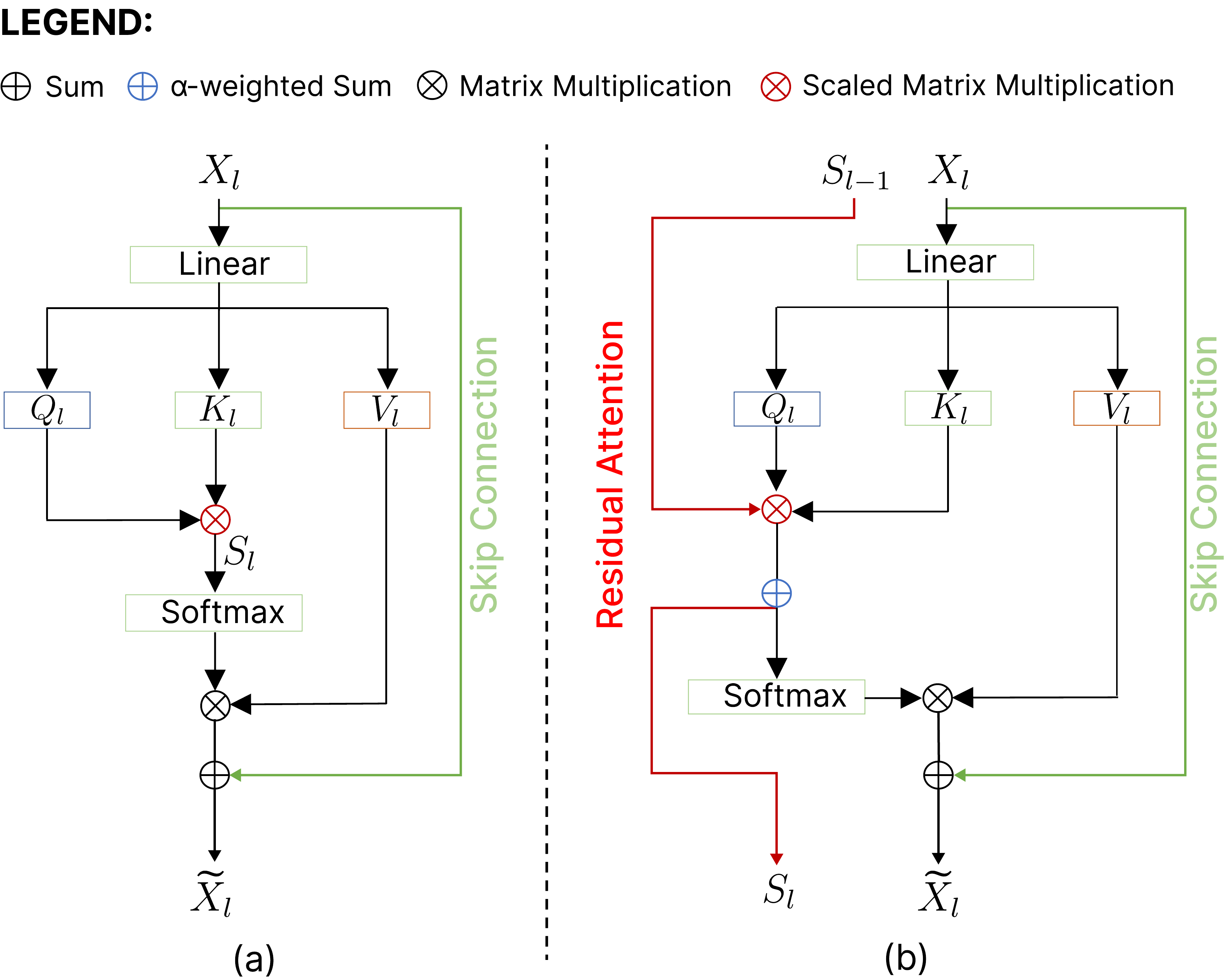} 
\caption{Illustration of the standard self-attention (a) and the proposed mechanism with residual attention module (b).}
\label{fig:attention}
\end{figure}
The standard ViT adopts the typical architecture as proposed in \cite{ashish@2017attention} with MLP and attention modules alternatively stacked throughout the network. Among different types of attention mechanisms, scaled dot-product self-attention is the most used one. It enables the model to extract complex relationships among the elements in the input data sequence and to dynamically assign varying significance degrees to each element according to the learned relationships \cite{ashish@2017attention}. Formally, given a set of image patch features (or tokens) $X = \{x_1, x_2, x_i,..., x_N\}$, with $X \in R^{N \times C}$, where $x_i$ is the vector representation of the $i$-th image token, $N$ is the total number of tokens, and $C$ is the channel size of tokens, a learnable linear projection layer is used to generate query $Q$, key $K$, and value $V$ vectors, such that $Q, K, V \in R^{N\times d}$ with $d$ being the projection dimensionality.
Once these vectors are obtained, for a given attention layer $l$, we first calculate the dot product between queries, scaling it by the square root of $d$ to prevent vanishing gradient \cite{ashish@2017attention} and produce raw attention scores vector $S_l$: 
\begin{equation} \label{eq:dot-prod}
    S_l = \frac{Q_lK^T_l}{\sqrt{d}},
\end{equation}
Therefore, the softmax function is applied to discretize such scores in the $[0,1]$ range, resulting in the attention weights matrix $A_l$:
\begin{equation} \label{eq:softmax}
    A_l = \textit{Softmax}(S_l).
\end{equation}
These weights determine how much each token contributes to the representation of other tokens. Finally, we compute the output features $\Tilde{X}_l$ by aggregating the value matrix $V_l$ using the attention weights:
\begin{equation} \label{eq:attention_output}
    \Tilde{X}_l = A_l V_l.
\end{equation}
%
%
Despite being exceptional in modeling token relationships, standard self-attention mechanisms can be computationally expensive when handling a large number of image patches or high-dimensional features. To tackle this, the MHSA approach, introduced in \cite{ashish@2017attention}, provides an elegant solution. Instead of processing the full-dimensional $Q$, $K$, and $V$ vectors directly, MHSA splits them into $H$ smaller subspaces named heads, each with a reduced dimension of $d/H$. This enables parallel calculations across these subspaces, allowing the model to simultaneously focus on different aspects of the input information. Formally, for a given head $h$ and layer $l$, the output attention features $\Tilde{X}_{l,h}$ can be obtained as follows:
\begin{equation} \label{eq:head_attention}
    \Tilde{X}_{l,h} = A_{l,h}V_{l,h},
\end{equation}
where $h$ represents the head index, $A_{l,h}$ and $V_{l,h}$ are the attention weights matrix and the value vectors of head $h$ and layer $l$, respectively. The outputs from each head are then combined, providing a comprehensive attention result that incorporates the insights gained from each individual subspace. Formally, according to the definition of MHSA, $\Tilde{X}_{l}$ can be represented as:
\begin{equation} \label{eq:mhsa}
    \Tilde{X}_{l} = Concat([A_{l,h}V_{l,h}]^H_{h=1}),
\end{equation}
where $Concat(\cdot)$ denotes the concatenation of the $H$ produced feature maps along the head dimension to provide the comprehensive MHSA results in $\Tilde{X}_{l}$.
The MLP module, on the other hand, extracts features independently from each patch and is usually constructed by stacking two linear layers with parameters $W_a$, $W_b$ respectively, and a non-linear activation function $\sigma$ in between them. This module also represents the last processing step of the transformer block, which takes in input $\Tilde{X}_l$ and produces the output features of layer $l$ denoted as $X_{l+1}$ since they are going to be used as the input features for the $l+1$ layer. Formally, it can be represented as:
\begin{equation}
    MLP(\Tilde{X}_l) = \sigma(W_{a,l} \Tilde{X}_l) W_{b,l},
\end{equation}
where $W_{a,l}$ and $W_{b,l}$ are the weights of the two stacked linear layers composing the MLP of layer $l$.

\subsection{Feature Collapsing} \label{ss:feat_collapse}
\begin{figure}
\centering
\includegraphics[width=0.6\textwidth]{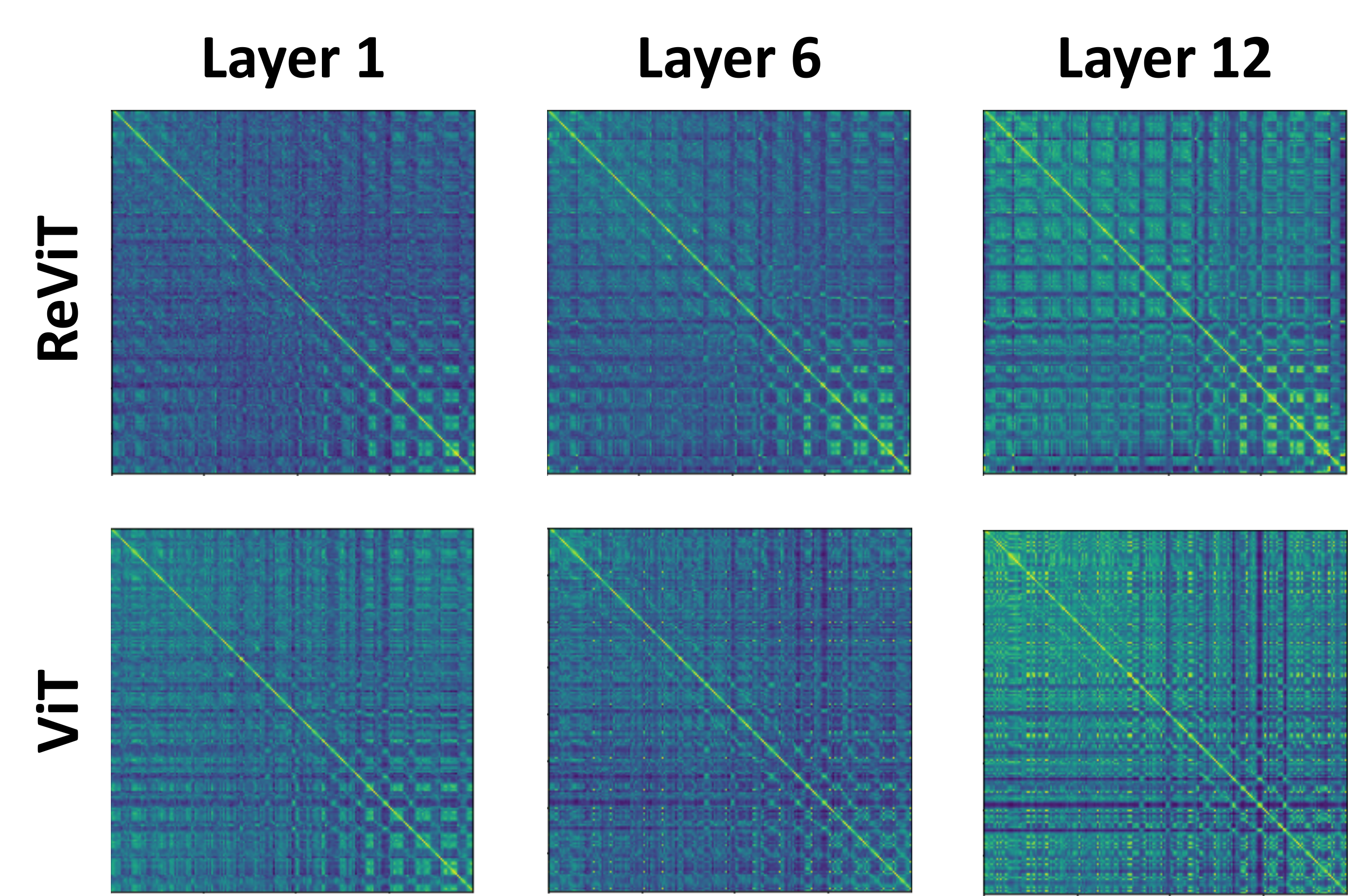}
 \caption{Feature similarity matrix in different layers for ViT and ReViT models. The color represents the similarity between each feature patch. A high frequency of green color shows high similarity between patches.}
 \label{fig:sim_matrix}
\end{figure}
Feature collapsing is a commonly observed occurrence within ViT architectures. It refers to the phenomenon where the features extracted from different image patches lose their distinctiveness and become increasingly similar or indistinguishable as the network depth grows. This phenomenon primarily arises due to the nature of the attention mechanism used in ViT, which progressively aggregates information between image patches as it traverses through the network layers. Such a process can be mathematically described by representing Eq. (\ref{eq:head_attention}) as a weighted summation of feature patches in the following manner:
\begin{equation}
   \Tilde{X}^i_{l,h} = \sum^N_{j=1} A^{i,j}_{l,h}V^j_{l,h}, \text{subject to} \sum^N_{j=1} A^{i,j}_{l,h} = 1, \text{for} \ i=[1, 2, ..., N].
\end{equation}
Here, $\Tilde{X}^i_{l,h}$ denotes the feature vector originating from patch $i$ and represents a weighted average of features from all patches j. The weights are determined by the values in the attention map $A_{l,h}$. Although the attention mechanism aims to capture global relationships between different feature patches \cite{dosovitskiy2021ViT}, it can lead to the loss of feature diversity, resulting in feature collapsing. In order to provide evidence for this observation, Fig. \ref{fig:sim_matrix} presents a visual representation of the feature similarity matrix for the ViT. This matrix is constructed by calculating the cosine similarity between distinct feature patches extracted by the ViT model. The key insight revealed by the figure is that, as the features progress from the shallower layers of the model to the deeper layers, there is a noticeable trend towards increased similarity among them.  

One approach to mitigate this phenomenon is the use of residual connections, which establish connections between features across different layers while bypassing the attention mechanism. Formally, the residual connection combined with MHSA operation can be expressed as follows:
\begin{equation}
    \text{residualMHSA}(X_l) = \text{MHSA}(X_l) + X_l.
\end{equation}
In this equation, the identity projection $X_l$ runs in parallel to the MHSA operation. Intuitively, because the features in $X_l$ exhibit greater diversity before passing through the MHSA operation, the summation of the two feature vectors preserves the characteristics from the previous layer. Consequently, the output retains more diverse features.  However, the residual connection between features ${X_l}$ proves insufficient. Empirical evidence supports this phenomenon in \cite{tang2021augmented}. 
Drawing inspiration from these discoveries, this study introduces a novel approach to tackling this issue. Specifically, it suggests an innovative alternative residual connection between attention layers (without bypassing the attention mechanism) designed to curb the rapid expansion of the attention across the head dimension, which is the main source of feature collapsing, and increase feature diversity as shown in Fig. \ref{fig:sim_matrix}.

\subsection{Residual Attention} \label{ss:res_attn}
\begin{figure}
\centering
\includegraphics[width=\textwidth]{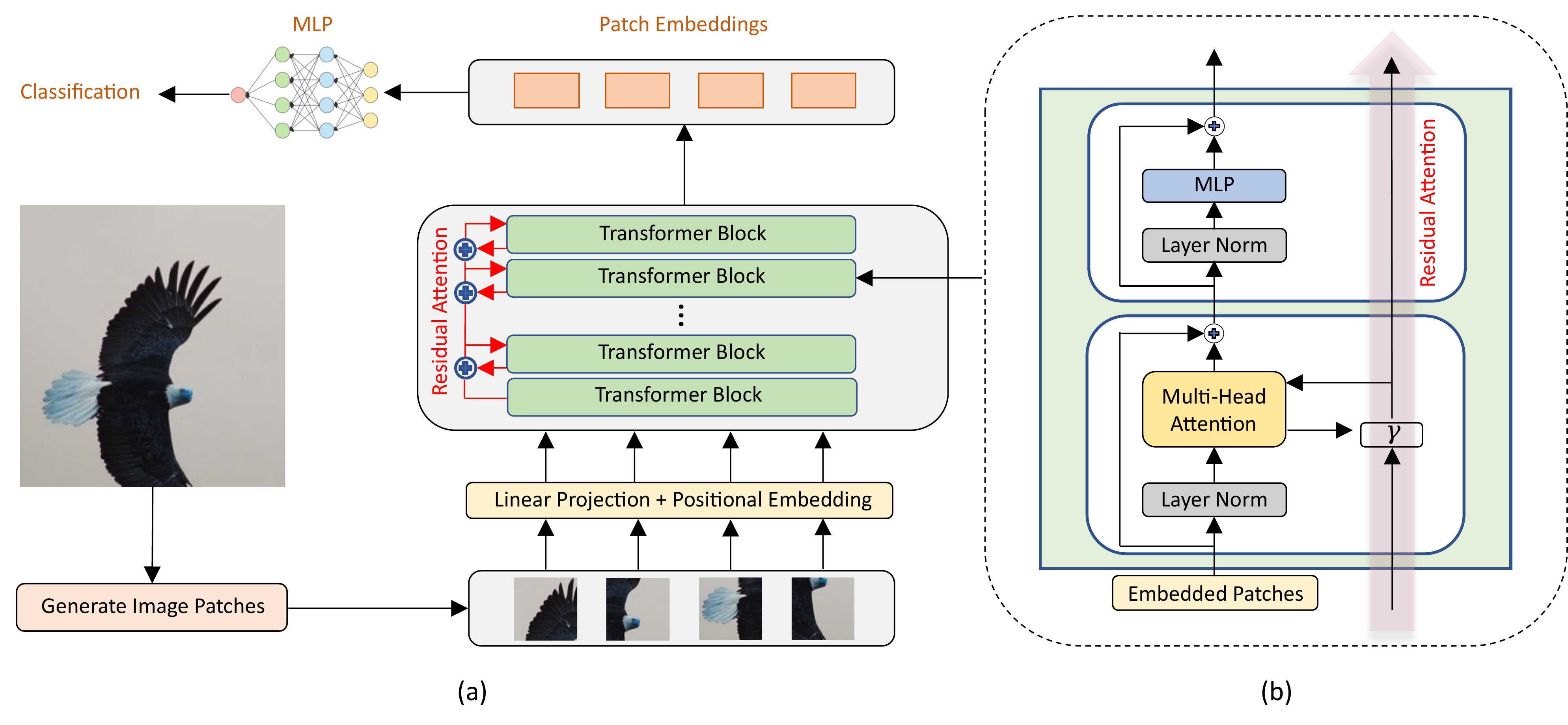}
 \caption{Illustration of the proposed ReViT network with residual attention. In (a) the overall architecture, in (b) the transformer block in detail, including the residual attention module.}
 \label{fig:ReViT_architecture}
\end{figure}
Generally, ViT model customization involves modifying the MHSA mechanism, normalization layers, MLP component, or residual connections among features, maintaining a specific information flow at the feature level between adjacent transformer blocks \cite{dosovitskiy2021ViT, Liu@2021swin}. Aiming to expand such a flow to improve the feature diversity, the ReViT architecture shown in Fig. \ref{fig:ReViT_architecture}(a) incorporates a novel skip connection between consecutive MHSA layers, enabling the propagation and accumulation of attention from shallow to deeper layers, as depicted in Fig. \ref{fig:ReViT_architecture}(b). This expansion, representing the proposed residual attention, is a complementary process to residual connections at the end of each transformer block. Indeed, while existing skip connections propagate the low-level features to deeper layers bypassing the MHSA layer, the residual attention propagates the information from $Q$ and $K$, which defines the relationships between patches used to extract these features, thus enabling a new attention capability to consider previously extracted relationships while learning to extract new ones. Formally, the proposed attention mechanism is implemented by changing the computation of the $S_l$ matrix defined in Eq. (\ref{eq:dot-prod}). Specifically, the equation is extended with an additional term representing the $Q$ and $K$ of the previous MHSA layers aggregated with those of the current layer, as follows:
\begin{equation}\label{eq:ra_scaled}
    S_{l} =
    \begin{cases}
     \frac{Q_{0} K_{0}^T}{\sqrt{d_0}}, & \text{if}\  l=0 \\
     \gamma(\frac{Q_{l} K_{l}^T}{\sqrt{d_l}}, S_{l-1}), & \text{otherwise}
    \end{cases},
\end{equation}
where $l$ indicates the $l$-th attention layer and $\gamma$ is the permutation invariant aggregation function. For $l=0$, the attention is computed as defined in Eq. (\ref{eq:dot-prod}). Otherwise, for $l > 0$, $S_{l}$ is obtained as the $\gamma$ of $S_{l-1}$ and the scaled dot product of $Q_{l}$ and $K_{l}$. Following this rationale, the information flow among adjacent MHSA layers is extended beyond the standard flow, which is limited only to feature forwarding, propagating the attention used to aggregate information between patches in previous layers. However, propagating and accumulating such information may prevent the network from learning high-level representations while amplifying the attention related to low-level features. To avoid this, $\gamma$ is implemented as a weighted sum where a learnable gate variable $\alpha$ is introduced to allow the network to autonomously determine how much attention to propagate between layers. Precisely, $\alpha$ balances the quantity of residual attention transmitted from the previous to the current MHSA layer during aggregation. Formally, the Eq. (\ref{eq:ra_scaled}) is further extended as follows:
\begin{equation}\label{eq:alpha_ra_scaled}
    S_{l} = 
    \begin{cases}
    \frac{Q_{l} \cdot K_{l}^T}{\sqrt{d_l}}, & \text{if}\  l=0 \\
    \alpha(\frac{Q_{l} \cdot K_{l}^T}{\sqrt{d_l}}) + (1 - \alpha)(S_{l-1}), & \text{otherwise}
    \end{cases},
\end{equation}
where $\alpha \in [0,1]$. According to this definition, for $l>0$, $S_{l}$ is obtained as a sum between the scaled dot product of ($Q_{l}$ $K_{l}$) and $S_{l-1}$ weighted by $\alpha$ and $1-\alpha$ respectively. After obtaining the balanced attention scores $S_{l}$, following Eq. (\ref{eq:softmax}), the softmax function is applied to compute the attention weights $A_l$ required for obtaining the new scaled dot-product attention output, as depicted in Fig. \ref{fig:attention}(b). Given the implementation of residual attention, it can be easily included in both existing single-scale or multi-scale vision transformer architectures, maintaining a similar computational cost.

\subsection{Attention Globality} \label{ss:attn_glob}
Attention globality is a concept that reflects the capacity of the attention mechanism to encompass and consider information from all patches across an image rather than being restricted to local or nearby elements. In other words, it represents the attention's global receptive field. This capability becomes particularly pronounced in the deeper layers of a ViT because these layers can extract high-level representations that account for long-range relationships between patches.
To quantify the globality of the attention mechanism across both heads and layers, researchers in \cite{d2021convit} introduce a metric known as the non-locality metric. This metric calculates, for each query patch $i$, the relative positional distances to all key patches $j$ weighted by their attention scores $A^{i,j}_{l,h}$. The resulting sum is then averaged by the number of patches to derive the non-locality metric for a specific head $h$. This can further be averaged across attention heads to obtain the non-locality metric for the entire layer. Mathematically, this metric is defined as follows:
\begin{equation}\label{eq:non-locality}
    D_{l,h} = \frac{1}{N} \sum_{i,j}A^{i,j}_{l,h}||\delta^{i,j}||,\\
    D_{l} = \frac{1}{H}\sum_h D_{l,h}.
\end{equation}
Here, $||\delta^{i,j}||$ represents the relative positional distance between query patch $i$ and key patch $j$, $D_{l,h}$ denotes the non-locality metric within a given layer $l$ and head $h$, while $D_{l}$ represents $D_{l,h}$ averaged along the head dimension. Moreover, since $||\delta^{i,j}||$ remains constant throughout the network (i.e., the relative positions of patches do not change), the value of $D_{l,h}$ relies entirely on $A_{l,h}$. Furthermore, as the non-locality metric $D$ increases in deeper layers (indicating a shift towards global attention), it suggests that the attention matrix $A_{l,h}$  (consequently $A_{l}$) in these layers generally has a greater capacity to capture global relationships across the different attention heads compared to attention matrix from the previous layer. This property of the attention mechanism can be harnessed to demonstrate the advantages of employing residual attention in slowing down the process of globalizing $A_l$. Taking into account Eq. (\ref{eq:alpha_ra_scaled}), which defines the calculation of $A_l$ with a residual connection, it considers attention scores not only from the current layer $l$ but also from previous layers. Consequently, the globality of layer $l$ in ReViT, at a given head, can be represented as:
\begin{equation}\label{eq:non-locality-ReViT}
    D_{l,h} = \frac{1}{N} \sum_{i,j}[\alpha(A^{i,j}_{l,h}) + (1-\alpha)(A^{i,j}_{l-1,h})]||\delta^{i,j}||.
\end{equation}
From a theoretical point of view, thanks to the above equation, given that the attention scores from earlier layers ($A_{l-1}$) generally tend to be more locally focused compared to those of the current layer $l$, the weighted combination of these sets of scores is expected to result in an attention matrix that exhibits less globality than if only the scores from the current layer were used. In other words, combining attention scores of given layer $l$ with those of previous layers through residual attention at each head $h$, slows down the globalization process. For empirical proof, refer to section \ref{ss:residual_attn_globality}.

%

\section{Experiments} \label{s:experiments}

This section details the implementation and evaluation of the proposed ReViT architecture. We conducted several experiments using well-established benchmark datasets for image classification, object detection, and instance segmentation to comprehensively assess the proposed architecture's effectiveness and generalization across various computer vision tasks in line with our goals. 

\subsection{Datasets and Performance Evaluation}

To rigorously evaluate ReViT's performance and versatility, we conduct an extensive empirical analysis across a diverse set of image classification benchmarks.  These benchmarks were chosen to encompass varying scales and complexities, ensuring a comprehensive assessment under various dataset scale scenarios. Specifically, we leverage ImageNet-1K, the de facto standard for image classification, to evaluate ReViT's ability to learn and generalize from a vast and diverse set of over 1 million images spanning 1,000 categories. In contrast, CIFAR-10 and CIFAR-100, with their smaller size and limited classes, allow us to assess ReViT's performance when training data is scarce, a common challenge in real-world applications.
Finally, we include the fine-grained datasets, Oxford Flowers-102 and Oxford-IIIT Pets, to scrutinize ReViT's robustness to image transformations and its capacity to discern subtle visual differences between visually similar classes. This evaluation is crucial for applications requiring precise identification.

To comprehensively assess ReViT's performance and generalization capabilities, we evaluate it on multiple fronts. First, we benchmark ReViT against standard ViT architectures and state-of-the-art image classification models using established metrics such as top-1 and top-5 accuracy. Next, we delve into ReViT's internal workings by quantitatively analyzing attention maps with the non-locality metric defined in Equations (\ref{eq:non-locality}-\ref{eq:non-locality-ReViT}) and qualitatively examining feature maps using GradCAM \cite{ramprasaath2017@gcam}. This analysis reveals how ReViT's attention mechanism contributes to its improved performance. Finally, we demonstrate ReViT's versatility by integrating it into multiscale transformer architectures (Swin and MViTv2). We assess its performance on object detection and instance segmentation tasks using the COCO2017 dataset, employing mean Average Precision (mAP) on bounding boxes and produced masks as the evaluation metrics, respectively. This multifaceted evaluation provides a holistic understanding of residual attention potential across diverse visual recognition tasks.

\subsection{Implementation Details} \label{ss:implementation}
The presented ReViT model is an augmented variant of the standard ViT architecture, tailored to enhance its performance by integrating the presented residual attention module. Specifically, we integrated the residual attention module on top of the standard ViT implementation by employing the PyTorch framework. Regarding the network version used in this study, we only relied on the base version of ViT containing 12 layers and denominated as ViT-B. Consequently, the augmented ViT is referred to as ReViT-B. To verify the versatility and effectiveness of ReViT, we also integrated the proposed approach to multiscale architectures, specifically the MViTv2 and Swin transformers. These experiments consider their respective tiny versions, i.e., MViTv2-T and Swin-T, each comprising 12 layers. The choice of these smaller-scale models, referred to as ReMViTv2-T and ReSwin-T in their augmented forms, was driven by a strategic aim to reduce computational load. This is particularly crucial as larger models typically necessitate extensive pre-training on vast datasets such as ImageNet21K. By focusing on these scaled-down versions, we aim to demonstrate that the proposed enhancements can deliver performance improvements even without the extensive computational resources typically required for larger networks. 

To train the augmented version of the networks, we follow the same settings as reported in each architecture's original study using an NVIDIA V100 GPU with 32GB of RAM. Specifically, for image classification on ImageNet1K, all models were trained for $300$ epochs, with an input image size of $224\times 224$, using gradient clipping and cosine scheduling learning rate warmup/decay. For the ReViT-B network training, Adam \cite{kingma@adam} was used as optimizer with a base learning rate set to $0.001$, $30$ warmup epochs, an effective batch size of $4096$, and the weight decay set to $0.3$. Instead, for ReSwin-T model training, AdamW optimizer \cite{loshchilov@2018adamW} was used by setting a base learning rate to $0.001$, $20$ warmup epochs, an effective batch size of $1024$, and the weight decay set to $0.05$. Finally, the remaining ReMViTv2-T network training was performed using the AdamW optimizer with a base learning rate set to $0.002$, $70$ warmup epochs, an effective batch size of $2048$, and the weight decay set to $0.1$. Regarding other image classification datasets, we use the same training setups reported in \cite{dosovitskiy2021ViT} for fair comparisons. Moreover, regarding the object detection and instance segmentation tasks, since such tasks require local awareness of features, the experiments were performed only using ReSwin-T and ReMViTv2-T. These models were trained for $36$ epochs using AdamW as an optimizer and linear scheduling learning rate decay. For ReSwin-T network training, the initial learning rate was set to $0.0001$, with a weight decay of $0.05$, and a batch size set to $16$. Instead, for ReMViTv2-T model training, the initial learning rate was set to $0.00016$, with a weight decay of $0.1$, and a batch size set to $64$. For more in-depth implementation details, we point the reader to our code repository \footnote{\href{https://github.com/ADiko1997/ReViT}{https://github.com/ADiko1997/ReViT}}.

\subsection{Image classification}
For the image classification tasks, models were primarily evaluated on ImageNet1K, considered as one of the most significant benchmarks. To evaluate the performance of the models, we used the top-1 single crop accuracy metric and reported the obtained results in Table \ref{tab:1}. As can be noticed, all models incorporating the residual attention module outperform their original network counterparts in the used metric. Specifically, ReViT-B, ReMViTv2-T, and ReSwin-T models with residual attention obtain an improvement of 4.6\%, 0.4\%, and 0.2\%, respectively, compared to their original counterparts. Notably, ReViT-B shows a significant improvement compared to ViT-B, indicating that residual attention is more effective in single-scale architectures that employ only global attention. This is because such a module was invented to contrast the over-globalization of the attention mechanism, a phenomenon manifested in the ViT-B architecture. On the contrary, multiscale architectures like Swin-T and MViTv2-T have their own in-built local mechanisms that help preserve the low-level features throughout the network. As such, the effect of the residual attention in these networks, i.e., ReSwin-T and ReMViTv2-T, is lower but still bolsters their representational ability and achieves improved performances compared to the original counterparts. Apart from the performance enhancements, another compelling factor highlighting the significance of residual attention within single-scale architectures is the parameter $\alpha$. Notably, in ReViT-B, $\alpha$ assumes a considerably smaller value when compared with the values observed in ReSwin-T and ReMViTv2-T. As Eq. (\ref{eq:alpha_ra_scaled}) indicates, this discrepancy signifies that prior attention holds substantially more significance in the context of ReViT. Additionally, Table \ref{tab:2} provides a more extensive comparison with the state of the art, showing that models with residual attention perform even more comparatively to the best models.

Subsequently, ReViT-B was tested on multiple image classification benchmarks, including CIFAR-10, CIFAR-100, Oxford Flowers-102, and Oxford-IIIT Pet, to simulate scenarios closer to real-world applications with data of small scale and low quality. Similar to ImageNet1K, top-1 accuracy on single crop images is used to assess the models. The performance of ReViT-B is then compared with that of the ViT-B model, and the results are presented in Table \ref{tab:3}. Notably, ReViT-B, with the residual attention, consistently outperforms ViT-B across all datasets with a minimum improvement of 0.6\% in CIFAR10 and a maximum of 3.4\% in CIFAR100.  This improvement can be attributed to the residual attention module's ability to capture a wider range of visual features, thus enhancing the network's capability to distinguish between different classes based on the image content. Note that the smaller improvement in CIFAR10 compared to CIFAR100 is due to the lower complexity of the former and the high accuracy of the standard ViT. The other models were not tested on these datasets because their official studies do not provide experimental results and focus only on large-scale datasets.

\begin{table}
    \centering
    \begin{tabular}{c|c|c|c|c|c}
         \textbf{Model} & \textbf{Image size} & \textbf{Params(M)} & \textbf{FLOPs(G)} & \textbf{top-1 acc. \%} & \textbf{$\alpha$} \\
         \hline \textbf{ViT-B} \cite{dosovitskiy2021ViT} & $224 \times 224$ & 86 & 17.5 & 77.8 & - \\
         \textbf{ReViT-B} & $224 \times 224$ & 86 & 17.5 & \textbf{82.4} & 0.56\\
         \hline \textbf{Swin-T} \cite{Liu@2021swin} & $224 \times 224$ & 29 & 4.5 & 81.3 & -\\
         \textbf{ReSwin-T} & $224 \times 224$ & 29 & 4.5 & \textbf{81.5} & 0.99\\
         \hline \textbf{MViTv2-T} \cite{Li@2022MViT} & $224 \times 224$ & 24 & 4.7 & 82.3 & - \\
         \textbf{ReMViTv2-T} & $224 \times 224$ & 24 & 4.7 & \textbf{82.7} & 0.99
    \end{tabular}
    \caption{Performance of ViT-based state-of-the-art models with and without residual attention on ImageNet1k single-crop top-1 accuracy. In addition, image size, model parameters in millions, i.e., Params(M), models throughput measured in theoretical floating point operations per second, i.e., FLOPs(G) \cite{dosovitskiy2021ViT}, and the final value of $\alpha$ are reported.}
    \label{tab:1}
\end{table}

\begin{table}
    \begin{tabular}{c|c|c|c}
         \textbf{Model}  & \textbf{Params(M)} & \textbf{FLOPs(G)} & \textbf{top-1 acc. \%} \\
         \hline \textbf{ViT-B} \cite{dosovitskiy2021ViT} & 86 & 17.5 & 77.8 \\
         \textbf{Deit-B} \cite{hugo@aug} & 86 & 17.5 & 81.8 \\
         \textbf{ConViT-B} \cite{d2021convit} & 86 & 17.5 & 82.4 \\
         \textbf{AugViT-B} \cite{tang2021augmented} & 86.5 & 17.5 & 82.4 \\
         \textbf{ReViT-B} & 86 & 17.5 & \textbf{82.4}\\
         \hline \textbf{Swin-T} \cite{Liu@2021swin} & 29 & 4.5 & 81.3\\
         \textbf{TNT-S} \cite{han2021transformer} & 24 & 5.2 & 81.5 \\
         \textbf{CoAtNet-0} \cite{dai2021coatnet} & 25 & 4.0 & 81.6 \\
         \textbf{T2T-ViT-14} \cite{yuan2021tokens} & 22 & 4.8 & 81.5 \\
         \textbf{T2T-ViT-14}\textsubscript{t} \cite{yuan2021tokens} & 22 & 6.1 & 81.7 \\
         \textbf{ConvNeXt-T} \cite{liu2022convnet} & 29 & 4.5 & 82.1 \\ 
         \textbf{MViTv2-T} \cite{Li@2022MViT} & 24 & 4.7 & 82.3 \\
         \textbf{VOLO-D1} \cite{yuan2022volo} & 27 & 7.1 & \textbf{84.2} \\
         \textbf{Dual-ViT-S} \cite{ting2023dualvit} & 24.6 & 5.4 & 83.4 \\
         \textbf{ReSwin-T} & 29 & 4.5 & 81.5\\        
         \textbf{ReMViTv2-T} &  24 & 4.7 & 82.7
         
    \end{tabular}
    \caption{Comparison with state-of-the-art models on ImageNet1K using input images of size $224 \times 224$.  In addition, information about model parameters in millions and throughput measured in floating point operations per second are reported.}
    \label{tab:2}
\end{table}

\begin{table}
    \centering
    \resizebox{\textwidth}{!}{
    \begin{tabular}{c|c|c|c|c|c}
         \textbf{Model} & \textbf{Image Size} & \textbf{CIFAR-10} & \textbf{CIFAR-100} & \textbf{Oxford-Flowers 102} & \textbf{Oxford-IIIT Pet} \\
         \hline ViT-B \cite{dosovitskiy2021ViT} & $384 \times 384$ & 98.1 & 87.1  & 89.5 & 93.8 \\
         ReViT-B &  $384 \times 384$ & \textbf{98.7} & \textbf{90.5} & \textbf{91.1} & \textbf{94.9}\\
    \end{tabular}}
    \caption{Top-1 accuracy performance comparison of ViT-B and ReViT-B on CIFAR-10, CIFAR-100, Oxford-Flowers 102, and Oxford-IIIT Pet datasets.}
    \label{tab:3}
\end{table}
\subsection{Translation Invariance} \label{ss:translation_invariance}
\begin{figure}
        \centering
	\subfloat[]{\includegraphics[width=0.33\textwidth]{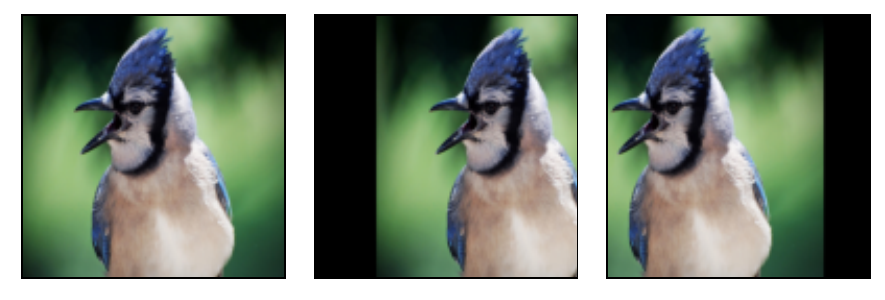}}\hfill 
	\subfloat[]{\includegraphics[width=0.33\textwidth]{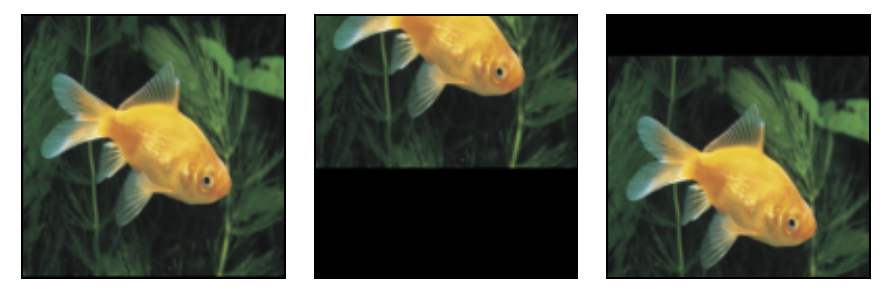}}\hfill
	\subfloat[]{\includegraphics[width=0.33\textwidth]{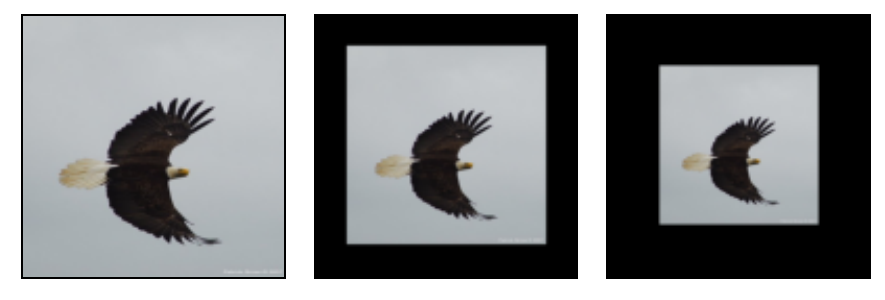}}
	\caption{Visualization of transformations showcasing translation invariance, specifically horizontal shifting (a), vertical shifting (b), and changes in scale (c). }\label{fig:invariance_transformations}
\end{figure}

In this study, the datasets used for visual recognition typically feature the object of interest centered within the image. This is in stark contrast to real-world scenarios where objects can be situated anywhere within the scene. To assess ReViT robustness, two experiments were devised to simulate varying scales and positions of the target object within the images as shown in Fig. \ref{fig:invariance_transformations}. The performance of ReViT-B and ViT-B in terms of top-1 accuracy on the Oxford Flowers-102 and Oxford-IIIT Pet datasets are compared in these experiments.

\begin{figure}
        \centering
	\subfloat[]{\includegraphics[width=0.33\textwidth]{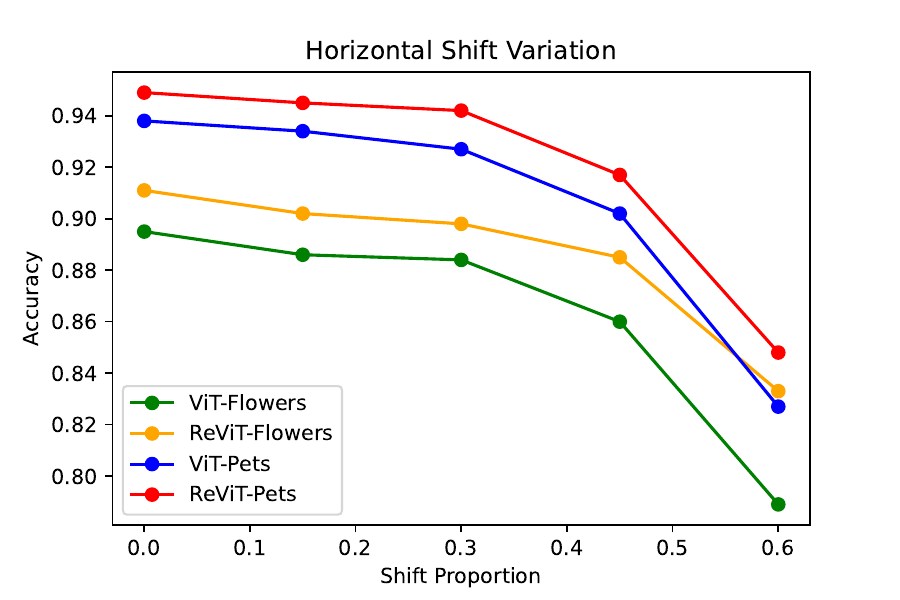}}\hfill 
	\subfloat[]{\includegraphics[width=0.33\textwidth]{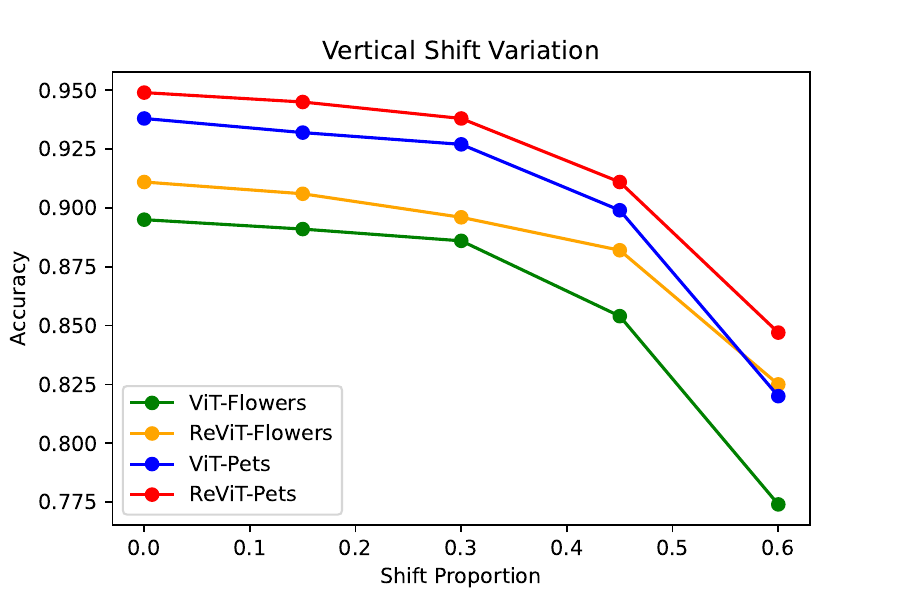}}\hfill
	\subfloat[]{\includegraphics[width=0.33\textwidth]{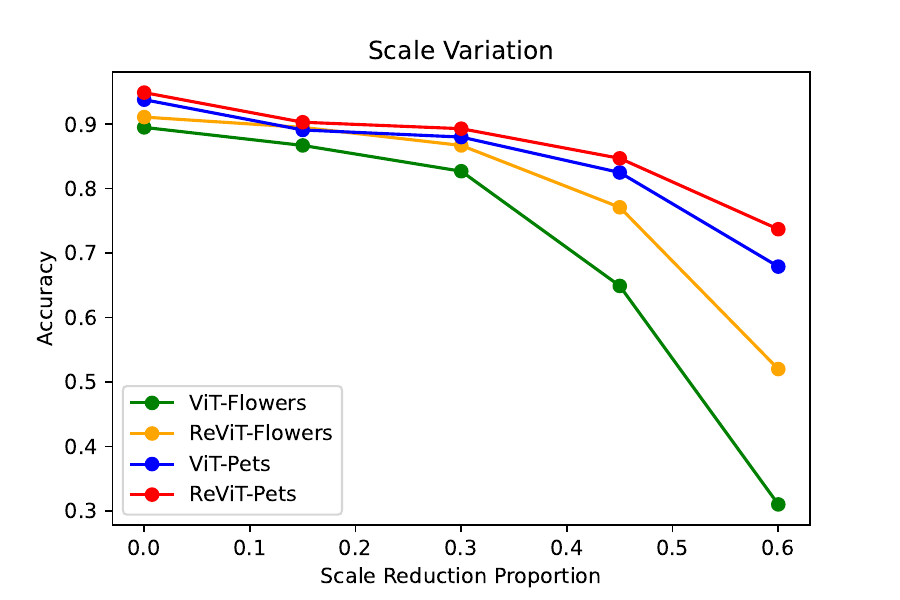}}
	\caption{Translation invariance test. The graphics depict how ViT-B and ReViT-B perform when subjected to variations in horizontal shifting (a), vertical shifting (b), and changes in size (c) on the Oxford Flowers-102 and Oxford-IIT Pet datasets.}\label{fig:translation_invariance}
\end{figure}

The first experiment evaluates the models' ability to handle horizontal and vertical translations. To achieve this, the input images are horizontally and vertically shifted on the image plane by \{15, 30, 45, 60\}\% of pixels, with the empty areas filled with zeros to maintain the same input image size as shown in Fig. \ref{fig:invariance_transformations}(a) and (b). As indicated in Table \ref{tab:5} and Fig. \ref{fig:translation_invariance}(a), both models exhibit a similar and impressive level of horizontal translation invariance, with only a minor reduction in performance. Overall, ReViT-B demonstrates slightly superior horizontal translation invariance, especially at the 60\% translation level. For the Oxford Flowers-102 dataset, ReViT-B performance decreases by 2.9\% less than ViT-B at a 60\% translation, while for the Oxford-IIIT Pet dataset, the decrease is 1.0\% less. Similar trends are observed for vertical translation invariance, as illustrated in Table \ref{tab:6} and Fig. \ref{fig:translation_invariance}(b). Although both models show a comparable drop in performance, ReViT-B displays better overall vertical translation invariance, particularly on the Oxford Flowers-102 dataset, where its performance decreases by 3.6\% less than ViT-B.

In the second experiment, the scale invariance of both models is evaluated, and the results are presented in Table \ref{tab:7} and Fig. \ref{fig:translation_invariance}(c). To generate images at different scales, the original images are resized by reducing their width and height by \{15, 30, 45, 60\}\%, with zero-padding to maintain the correct input size as illustrated in Fig. \ref{fig:invariance_transformations}(c). Unlike the first experiment, ReViT-B consistently outperforms ViT-B in terms of scale invariance. Most notably, on the Oxford Flowers-102.
Particularly, this dataset exhibits high intra-class variation and inter-class similarity for certain categories. Thus, accurately identifying flower categories requires capturing fine details like shape, color, and proportions. When images are downscaled significantly (beyond 45\%), the details become much smaller and the number of informative tokens in the sequence decreases. Additionally, the effective distance between informative tokens shrinks as patched regions (filled with zeros) do not affect the attention calculations (i.e., multiplication with 0, no information to aggregate). This reduced distance allows attention to rapidly reach global peaks, leading to earlier feature collapse.
Under these conditions, the vanilla ViT, lacking an anti-collapse mechanism, experiences a significant performance drop of up to 58.5 percentage points. This is 19.5 percentage points or 33.3\% more than ReViT's reduction. The key difference lies in ReViT's residual attention mechanism, which counteracts feature collapsing.
These results highlight that ReViT-B possesses significantly better scale invariance compared to ViT-B.

\begin{table}
    \setlength{\tabcolsep}{4pt}
    \centering
    \begin{tabular}{c|c c c c| c c c c}
         \multicolumn{9}{c}{Horizontal Translation} \\
         \hline & \multicolumn{4}{c|}{\textbf{Oxford Flowers-102}} & \multicolumn{4}{c}{\textbf{Oxford-IIIT Pet}} \\
         \textbf{Model} & \textbf{15\%} & \textbf{30\%} & \textbf{45\%} & \textbf{60\%} & \textbf{15\%} & \textbf{30\%} & \textbf{45\%} & \textbf{60\%} \\
         \hline \textbf{ViT-B} & 0.9 & \textbf{1.1} & 3.5 & 10.6 & 0.4 & 1.1 & 3.6 & 11.1\\
         \textbf{ReVit-B} & \textbf{0.8} & 1.2 & \textbf{2.5} & \textbf{7.7} & 0.4 & \textbf{0.7} & \textbf{3.2} & \textbf{10.1} \\
         \hline $\Delta$ & 0.1 & -0.1 & 1.0 & 2.9 & 0.0 & 0.4 & 0.4 & 1.0
    \end{tabular}
    \caption{Reduction in performance (in \%) observed when applying a horizontal translation to input images in the Oxford Flowers-102 and Oxford-IIIT Pet datasets. $\Delta$ represents the variance in the extent of performance degradation between ReViT-B and ViT-B.}
    \label{tab:5}
\end{table}
\begin{table}
    \setlength{\tabcolsep}{4pt}
    \centering
    \begin{tabular}{c|c c c c| c c c c}
         \multicolumn{9}{c}{Vertical Translation} \\
         \hline & \multicolumn{4}{c|}{\textbf{Oxford Flowers-102}} & \multicolumn{4}{c}{\textbf{Oxford-IIIT Pet}} \\
         \textbf{Model} & \textbf{15\%} & \textbf{30\%} & \textbf{45\%} & \textbf{60\%} & \textbf{15\%} & \textbf{30\%} & \textbf{45\%} & \textbf{60\%} \\
         \hline \textbf{ViT-B} & 0.4 & \textbf{0.9} & 4.1 & 12.1 & 0.6 & 1.1 & 3.9 & 11.8\\
         \textbf{ReVit-B} &  0.4 & 1.4 & \textbf{3.0} & \textbf{8.5} & \textbf{0.4} & 1.1 & \textbf{3.8} & \textbf{10.2} \\
         \hline $\Delta$ & 0.0 & -0.5 & 1.1 & 3.6 & 0.2 & 0.0 & 0.1 & 1.6
    \end{tabular}
    \caption{Diminished performance (in \%) resulting from a vertical translation applied to input images within the Oxford Flowers-102 and Oxford-IIIT Pet datasets. The symbol $\Delta$ indicates the difference in performance reduction between ReViT-B and ViT-B.}
    \label{tab:6}
\end{table}
\begin{table}
    \setlength{\tabcolsep}{4pt}
    \centering
    \begin{tabular}{c|c c c c| c c c c}
         \multicolumn{9}{c}{Scale Variance} \\
         \hline & \multicolumn{4}{c|}{\textbf{Oxford Flowers-102}} & \multicolumn{4}{c}{\textbf{Oxford-IIIT Pet}} \\
         \textbf{Model} & \textbf{15\%} & \textbf{30\%} & \textbf{45\%} & \textbf{60\%} & \textbf{15\%} & \textbf{30\%} & \textbf{45\%} & \textbf{60\%} \\
         \hline \textbf{ViT-B} & 2.8 & 6.8 & 24.6 & 58.5 & 0.4 & 1.5 & 7.0 & 25.9\\
         \textbf{ReVit-B} &  \textbf{1.5} & \textbf{4.3} & \textbf{13.9} & \textbf{39.0} & 0.4 & \textbf{1.4} & \textbf{6.0} & \textbf{21.2} \\
         \hline \textbf{$\Delta$} & 1.3 & 2.5 & 9.7 & 19.5 & 0.0 & 0.1 & 1.0 & 4.7
    \end{tabular}
    \caption{Diminished performance (in \%) resulting from variations in scale applied to input images within the Oxford Flowers-102 and Oxford-IIIT Pet datasets. $\Delta$ indicates the disparity in performance decline between ReViT-B and ViT-B.}
    \label{tab:7}
\end{table}

\subsection{Residual Attention and Globality} \label{ss:residual_attn_globality}
\begin{figure}[ht]
        \centering
	\subfloat[]{\includegraphics[width=0.33\textwidth]{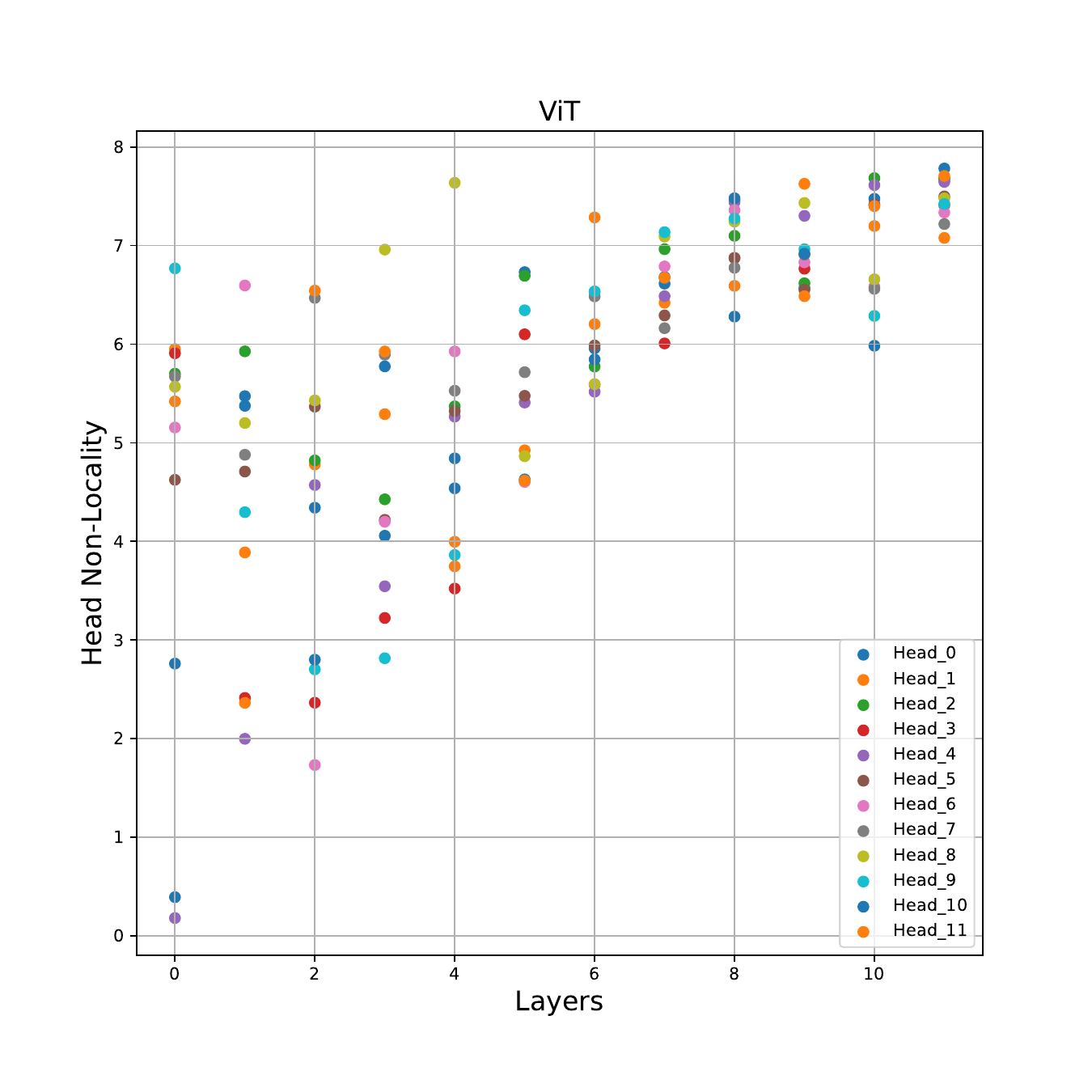}}\hfill 
	\subfloat[]{\includegraphics[width=0.33\textwidth]{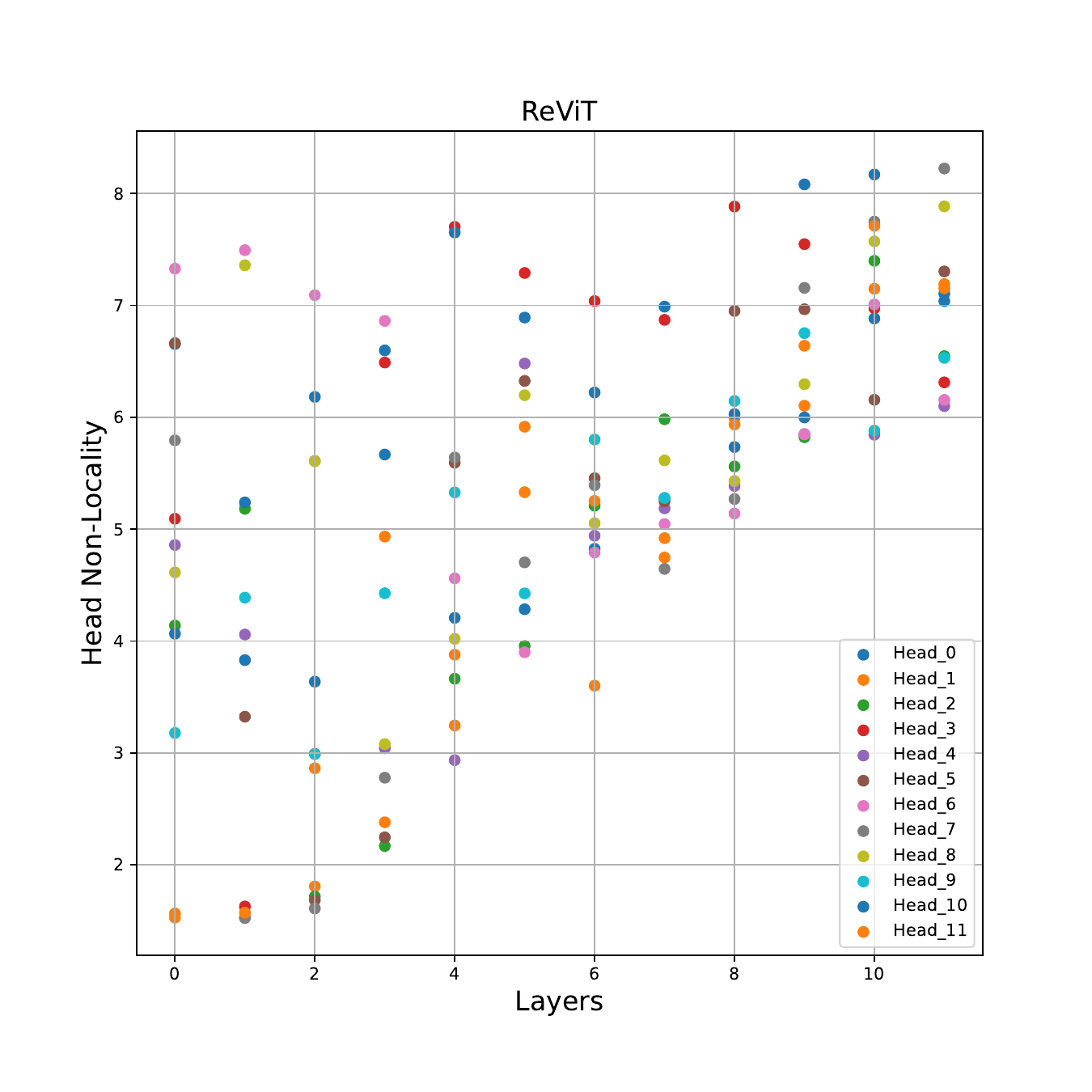}}\hfill
	\subfloat[]{\includegraphics[width=0.32\textwidth]{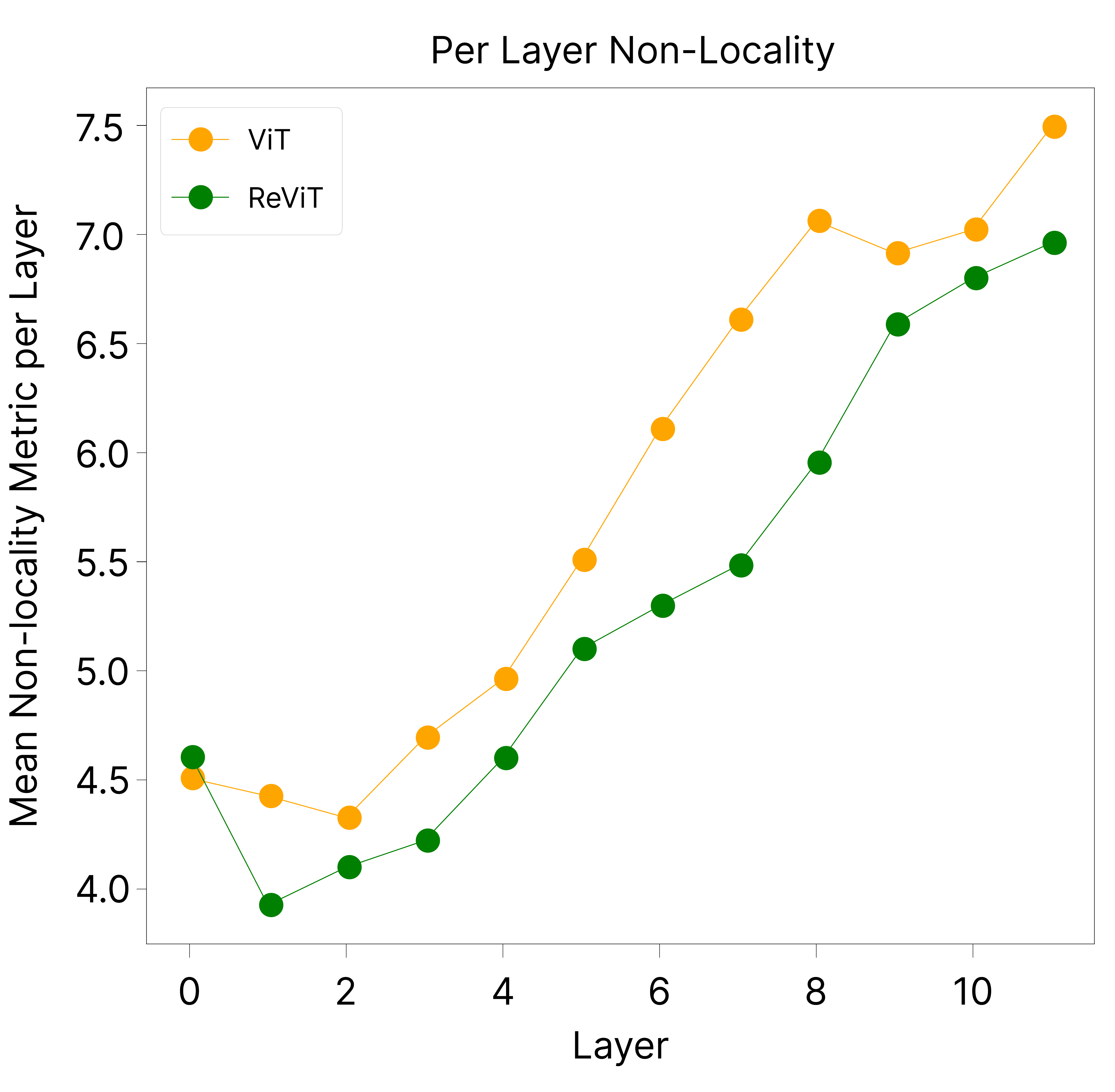}}
	\caption{The non-locality metric is depicted for individual heads across all layers in the case of ViT-B in (a) and ReViT-B in (b). This visualization illustrates how the receptive field of each head expands as it progresses through deeper layers. On the other hand, in (c), a comparison of the non-locality metric between ViT-B and ReViT-B for each layer is illustrated.}\label{fig:ViT_vs_ReViT}
\end{figure}
The investigation into the global reach of MHSA within vision transformers, both with and without residual attention, aims to examine the hypothesis that residual attention decelerates the globalization of the attention mechanism. To assess this hypothesis quantitatively, we employ the non-locality metric calculated according to Eq. (\ref{eq:non-locality}). Additionally, the metric is computed at both the head and layer levels. The results of this analysis, depicted in Fig. \ref{fig:ViT_vs_ReViT}, display the average metric across 256 input images from the ImageNet1K validation set for ViT-B and ReViT-B models trained on the same dataset. The x-axis on the plots corresponds to the layer index, while the mean non-locality for each head or layer is represented on the y-axis. Fig. \ref{fig:ViT_vs_ReViT}(a) illustrates that in ViT-B, the non-locality of each head increases as it delves deeper into the layers, approaching globality which corresponds to the findings in \cite{dosovitskiy2021ViT, d2021convit}. A similar trend is observed in ReViT-B, as seen in Fig. \ref{fig:ViT_vs_ReViT}(b), albeit with some noteworthy distinctions. Notably, ViT-B exhibits a more consistent and monotonic increase in non-locality, with minimal variations between different heads, particularly after layer 5. In contrast, ReViT-B displays more diverse differences between heads, with some heads finding difficulty in escaping the locality while others become completely global, resulting in a less monotonic increase in non-locality. Such behavior is caused by the residual attention mechanism, which induces locality from previous layers into the attention of current layers. In summary, the results substantiate the hypothesis that the attention receptive field of heads in vision transformers becomes global in deeper layers, a phenomenon observed in both ViT-B and ReViT-B models. However, these findings also reveal that ReViT-B retains the ability to extract global relationships in deeper layers while still maintaining some heads that remain localized with relatively low non-locality values. Simultaneously, ViT-B is confined solely to global relationships, leading to feature collapse. To provide a broader perspective, Fig. \ref{fig:ViT_vs_ReViT}(c) presents the non-locality of ViT-B and ReViT-B at the layer level, represented as the average non-locality metric across all heads within each layer. It is evident that when computed using the same 256 input examples, ReViT-B consistently maintains lower non-locality values (except the first), indicating its ability to incorporate low-level features, with the help of relatively local heads, into the representations learned from heads that exhibit non-locality escape (i.e., become global).
\begin{figure}[ht]
\centering
\includegraphics[width=\textwidth]{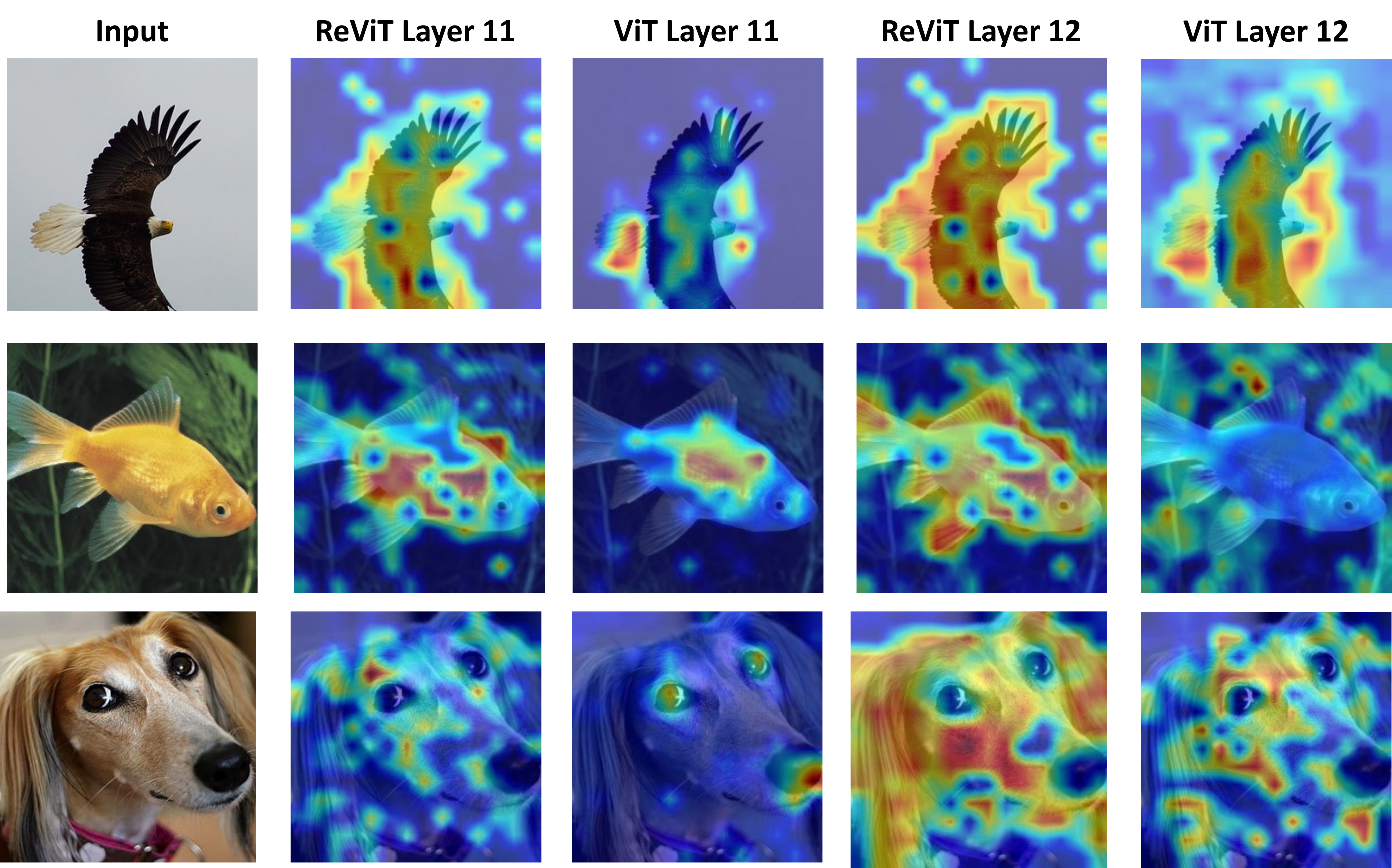}
 \caption{Comparison of feature maps in the last two layers of ViT-B and ReViT-B extracted using GradCAM algorithm from the illustrated input images.}
 \label{fig:gradcam}
\end{figure}

\subsection{Qualitative Comparisons}\label{ss:qualitative_comparison}
The effectiveness of the proposed residual attention module is further demonstrated through a qualitative evaluation of the features learned by ViT-B and ReViT-B using the MHSA mechanism, both with and without residual attention, respectively. To enable such a comparison, we employ the GradCAM algorithm and apply it to the last two MHSA layers of ViT-B and ReViT-B trained on ImageNet1k with samples from the validation set of the same dataset. It's worth emphasizing that layers 11 and 12 were selected due to their high globality, aligning with the objective of this experiment, which is to highlight how ReViT-B incorporates low-level features into its learned representations compared to ViT-B. To proceed, the feature maps of both models and the input images used to obtain such features are illustrated in Fig. \ref{fig:gradcam}. As can be noticed, the feature maps extracted from ViT-B are poor in detail and lack low-level features like shapes and edges. This is caused by the globalization of its attention mechanism, which provokes the feature-collapsing phenomenon. This also proves the observation made by authors in \cite{d2021convit}, which states that ViT-based architectures relying on the standard MHSA often exhibit limited attention variability, resulting in reduced feature diversity or even feature collapse \cite{tang2021augmented}. On the contrary, ReViT-B, equipped with residual attention, displays greater feature diversity with feature maps almost completely aligned with the entire region where the object of interest is located, preserving the same shape. This phenomenon suggests that ReViT-B can incorporate low-level information like shapes and edges with the global context of the scene. Additionally, such ability can be attributed to residual attention. Thanks to this mechanism, the network can propagate attention maps from shallow layers to deeper layers, slowing down the globalization of attention, improving feature diversity, and incorporating low-level characteristics into the global context. Notably, this augmented feature diversity aligns with ReViT's improved performance in terms of image classification accuracy compared to the standard ViT architecture and its ability to extract important low-level information in the presence of translation invariances.

\subsection{Object Detection and Instance Segmentation}
To assess the general applicability of the residual attention module, we test its knowledge transfer ability in downstream tasks. As such, further experiments were conducted on spatial-aware networks, i.e., ReSwin-T and ReMViTv2-T, for the tasks of object detection and instance segmentation, achieving increased results. The evaluation was performed on the COCO2017 dataset for both tasks using the COCO average precision (AP) metric, and the input images were resized with $480 \leq H \leq 800$ and $ W = 1333$. For fair comparisons with the solutions proposed in \cite{Liu@2021swin} and \cite{Li@2022MViT}, the Cascade Mask-RCNN \cite{cascade} and the Mask R-CNN \cite{he@2017mask} models were used with the ReSwin-T and ReMViTv2-T as backbones, respectively. Additionally, both ReSwin-T and ReMViTv2-T were initialized with the weights learned during ImageNet1K training, hence, knowledge transferring. Table \ref{tab:od_seg} summarizes results on the subset of 5000 validation images. For the object detection task, the ReSwin-T model performance is higher for $AP_{box}$ and $AP_{box}^{50}$ with respect to the original Swin-T architecture. Regarding the ReMViTv2-T model, the performance improves for all the considered APs compared to the original MViTv2-T architecture. For the instance segmentation task, ReSwin-T and ReMViTv2-T with residual attention increase the overall performance in all the metrics compared to their original works, even though by a small margin. Additionally, even though residual attention boosts the performances of such models, its impact is lower compared to single-scale architectures due to the multi-scale nature of such architectures and the presence of local information in the produced features. 

\begin{table}
    \setlength{\tabcolsep}{3pt}
    \centering
    \begin{tabular}{c|c|c c c|c c c|c}
         \textbf{Framework} & \textbf{Backbone} & \textbf{$AP_{box}$} & \textbf{$AP_{box}^{50}$} & \textbf{$AP_{box}^{75}$} & \textbf{$AP_{mask}$} & \textbf{$AP_{mask}^{50}$} & \textbf{$AP_{mask}^{75}$} & \textbf{\#Params} \\
         
         \hline Cascade Mask & Swin-T \cite{Liu@2021swin} & 50.5 & 69.3 & 54.9 & 43.7 & 66.6 & 47.1 & 86M \\
         
          R-CNN & ReSwin-T & \textbf{50.6} & \textbf{69.7} & 54.9 & \textbf{43.9} & \textbf{66.8} & \textbf{47.6} & 86M \\
          
         \hline Mask & MViTv2-T \cite{Li@2022MViT} & 48.2 & 70.9 & 53.3 & 43.8 & 67.9 & 47.2 & 44M \\
         
        R-CNN & ReMViTv2-T & \textbf{48.5} & \textbf{71.0} & \textbf{53.5} & \textbf{44.0} & \textbf{68.3} & \textbf{47.3} & 44M \\
    \end{tabular}
    \caption{Performance of the residual attention module implemented on top of Swin and MViTv2 architectures for object detection and instance segmentation on COCO2017 $5K$ validation set.}
    \label{tab:od_seg}
\end{table}

%
%
%
%
%

\section{Ablation Study}
In this section, we perform an ablation study to determine the significance of balancing the past and current information in the proposed residual attention and its effect when is specific to each layer. We will specifically examine the impact of $\alpha$ on various image classification benchmarks, including CIFAR10, CIFAR100, Oxford Flowers-102, and Oxford-IIIT Pet.
\subsection{Inspecting the effect of $\alpha$}
\begin{figure}[ht]
        \centering
	\includegraphics[width=0.5\textwidth]{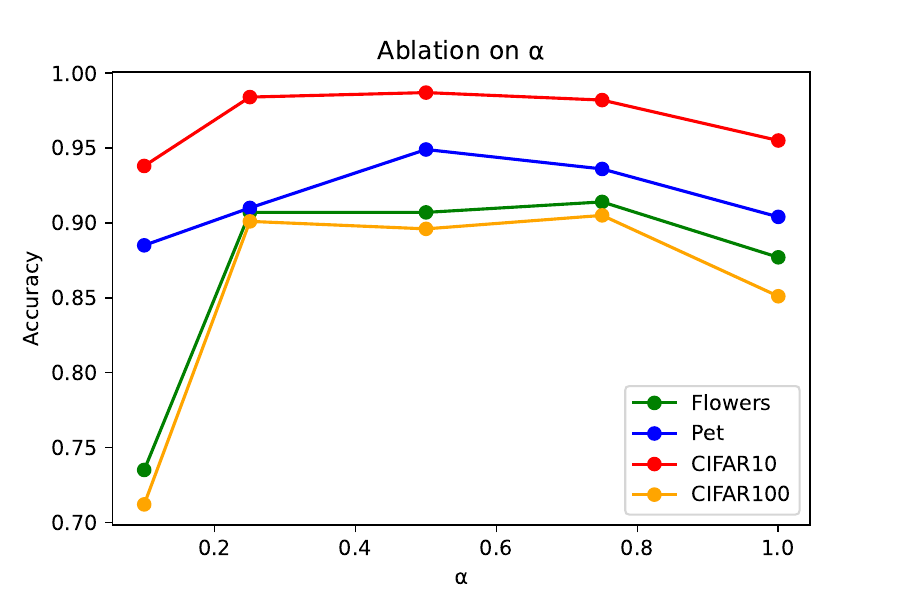}	
 \caption{ReViT classification performance on CIFAR10, CIFAR100, Oxford Flowers-102 and Oxford-IIIT Pet datasets for different values of $\alpha$.}
 \label{fig:ablation_on_alpha}
\end{figure}
The notion of residual attention acts as a conduit for propagating attention information through diverse layers via a shortcut connection, formally represented in Eq. (\ref{eq:alpha_ra_scaled}). Introducing a trainable gate variable, denoted as $\alpha$, ensures a balanced incorporation of residual attention information within this connection. The pivotal role of $\alpha$ lies in its ability to regulate past and present information flow, maintaining a harmonious equilibrium. A meticulous series of experiments are devised to study this variable's impact empirically.

These experiments encompass an assessment of ReViT's performance in classification tasks, with the gating variable held constant while manually adjusting its values to observe the model's behavior. Specifically, ReViT is tested across four classification benchmarks: CIFAR10, CIFAR100, Oxford Flowers-102, and Oxford-IIIT Pet, with $\alpha$ values spanning from 0 to 1. With 0 being the lower extremity considering only past attention and 1 being the other extremity completely ignoring past attention (i.e., ViT). The outcomes of these experiments are graphically presented in Fig. \ref{fig:ablation_on_alpha}. For $\alpha=0$, according to Eq. (\ref{eq:alpha_ra_scaled}), ReViT relies only on the attention information from the first layer, and as expected, the performance decreases, especially on complex datasets like CIFAR100 and Oxford Flowers-102. Such performance collapse is explained by the fact that for an $\alpha=0$, ReViT lacks global context. For the other values of $\alpha$, notably in datasets characterized by a modest number of classes and a surplus of examples per class, such as CIFAR10, the impact is relatively subdued. However, in a more demanding dataset featuring significant inter-class and intra-class variations compared to CIFAR10, like Oxford-IIIT Pet, the role of $\alpha$ becomes more pivotal, exemplified by the model's sensitivity to varying $\alpha$ values affecting its performance. Furthermore, in datasets like CIFAR100 and Oxford Flowers-102, which are even more complex, the effect of $\alpha$ on ReViT becomes even more important and manifests quite similarly with the model attaining peak performance at $\alpha=0.75$, and experiencing a sharp decline as it approaches 1. This increased importance of $\alpha$ on complex dataset is related to the higher necessity of low-level features like colors and shapes needed to discriminate between different classes of such dataset characterized by high inter-class similarity and intra-class variances. Finally, considering that an $\alpha$ value close to 1 typically does not yield the optimal performance across all datasets, it becomes evident that past attention information is crucial for learning better representation, and $\alpha$ assumes a vital role within ReViT. It balances the integration of past and current attention information, facilitating the extraction of high-quality image representations encompassing high and low-level features.

\subsection{Learning $\alpha$ for each layer}
\begin{figure}[ht]
        \centering
	\subfloat[]{\includegraphics[width=0.49\textwidth]{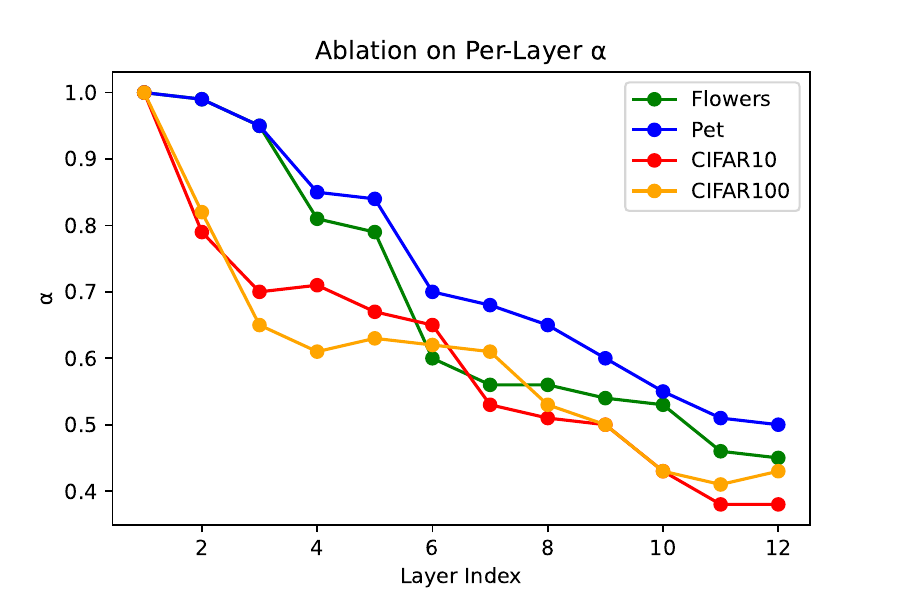}}
	\subfloat[]{\includegraphics[width=0.49\textwidth]{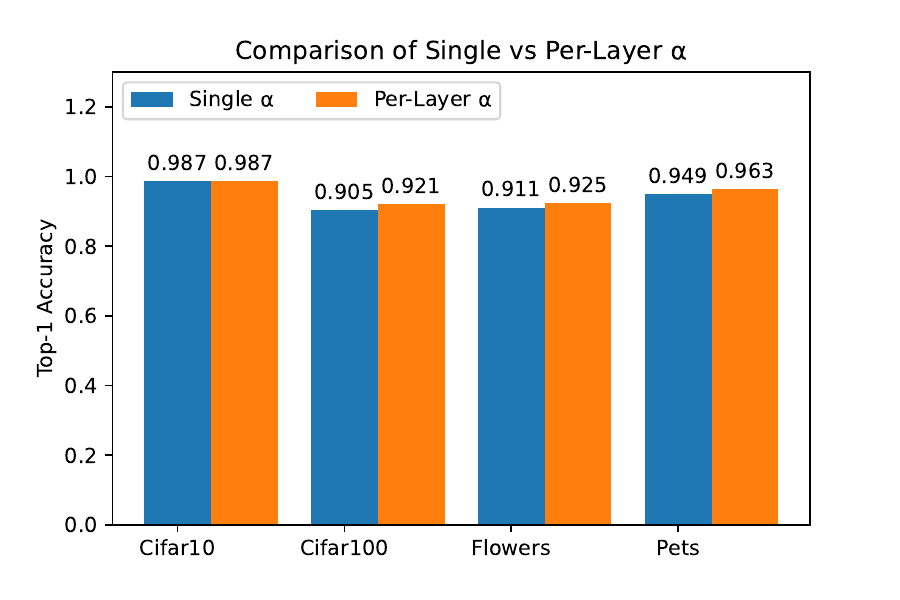}}
 
	\caption{The effect of learning layer-specific $\alpha$ values. In (a), the $\alpha_l$ values for each layer of ReViT-B on the chosen datasets are reported. In (b), the impact of layers-specific $\alpha$ on the accuracy of ReViT-B on the same datasets is reported.}\label{fig:per_layer_alpha}
\end{figure}

According to Eq. (\ref{eq:alpha_ra_scaled}), alpha is a learnable weight scalar that tries to balance the attention flow between consecutive blocks. Additionally, $\alpha$ is kept the same across all layers (i.e., the weight of current and past attention is the same at each layer). Such design choice was intended to resemble a traditional residual connection to show that even a simple operation such as a weighted sum can help with feature diversity when applied to the main catalyst: self-attention. As this operation has shown its benefits, we now give each layer the freedom to choose its balance between past and current attention by making $\alpha$ layer-specific. Mathematically, this means that Eq.(\ref{eq:alpha_ra_scaled}) is slightly changed as following:
\begin{equation}\label{eq:alpha_ra_scaled_v2}
    S_{l} = 
    \begin{cases}
    \frac{Q_{l} \cdot K_{l}^T}{\sqrt{d_l}}, & \text{if}\  l=0 \\
    \alpha_l(\frac{Q_{l} \cdot K_{l}^T}{\sqrt{d_l}}) + (1 - \alpha_l)(S_{l-1}), & \text{otherwise}
    \end{cases},
\end{equation}
Notably, as the computational logic remains the same as in Eq. (\ref{eq:alpha_ra_scaled}), the subscript $l$ on $\alpha$ (i.e., $\alpha_l$) indicates that $\alpha$ is bounded to the specific layer. To proceed with analysing its effect, we train ReViT-B with residual attention implemented according to Eq. (\ref{eq:alpha_ra_scaled_v2}) on CIFAR10, CIFAR100, Oxford Flowers-102, and Oxford-IIIT Pet datasets and report the per-layer $\alpha_l$ value and its impact on the overall performance compared to the default configuration in Fig. \ref{fig:per_layer_alpha} (a) and (b) respectively. 

Upon carefully analyzing the values of $\alpha_l$ for different layers as depicted in Fig. \ref{fig:per_layer_alpha} (a), we observe a trend where $\alpha_l$ decreases as information flows deeper into the layers, placing greater emphasis on past attention information. This behavior aligns with the intended design of $\alpha_l$, considering how feature diversity evolves as information progresses towards deeper layers. Specifically, based on our analysis from previous sections, we found that the diversity of features is higher in the top layers of the network. Such observation is further supported by the fact that $\alpha_l$ is high in such layers,  reducing the necessity of using past attention. In contrast, beyond the fifth layer, where feature diversity diminishes significantly, $\alpha_l$ drops drastically, highlighting the importance of the attention from previous layers derived from the interactions of tokens with higher feature diversity. Moreover, we also notice that for datasets with higher image quality, such as Oxford Flowers-102 and Oxford-IIIT Pet, $\alpha_l$ decreases slower in the first half of the layers compared to CIFAR10 and CIFAR100 which are known to be low-quality. Such a phenomenon might suggest that in higher-quality images, features tend to lose diversity slower in the shallow layers but still collapse similarly in the deeper ones. Finally, in Fig. \ref{fig:per_layer_alpha} (b), we compare the performance of ReViT-B on the four datasets with single learnable $\alpha$ and per layer learnable $\alpha_l$. From the results, we conclude that there is an improvement in performance from having learnable $\alpha_l$, except in the case of the CIFAR10 dataset. Specifically, for CIFAR10, the obtained results are identical to the setting with global $\alpha$; instead, for CIFAR100, Oxford Flowers-102, and Oxford-IIIT Pet, improvements of 0.16, 0.14 and 0.14, respectively, are obtained by using a per-layer $\alpha$. However, it is worth noting that the margin for improvements in these datasets is low, especially on CIFAR10, with accuracy going beyond 98\%.

\section{Conclusion}

In this work, we have contributed the \textbf{Re}idual attention \textbf{Vi}sion \textbf{T}ransformer (ReViT) to the family of vision transformers, designed for contrasting the phenomenon of feature collapsing. While previous works approach the targeted problem by augmenting feature connectivity \cite{hugo@aug}, inducing convolution-like bias \cite{d2021convit} or shaping the transformer architecture to resemble a CNN \cite{Liu@2021swin,Li@2022MViT}, ReViT offers a more elegant, simple, target focused and effective solution by addressing the over globalization of the attention mechanism through residual attention connections. These connections give ReViT the ability to propagate attention information across layers, allowing it to capture and retain fine-grained details that are crucial for accurate visual recognition. Extensive experiments on various recognition benchmarks, coupled with in-depth analysis of the residual attention mechanism, showcase ReViT's effectiveness across diverse tasks and its ability to preserve feature diversity in deeper transformer layers.

The practical implications of ReViT's enhanced feature diversity are far-reaching. In domains such as medical imaging or specialized industrial inspection, where obtaining large labeled datasets is often infeasible, ReViT offers a promising solution. Its ability to learn effectively even in limited data and capture a rich array of features has the potential to unlock the utility of standard transformer architectures even in such scenarios \cite{li2022locality}. Another key strength of ReViT lies in its simple yet effective design. The attention residual connection seamlessly integrates into existing ViT blocks, incurring minimal to no computational overhead while yielding substantial performance gains. This architectural simplicity preserves the core fundamentals of the standard transformer architecture, aligning with the future prospective of designing unified software and hardware solutions given the ability of such architectures to generalize across tasks, domains, and modalities \cite{girdhar2023imagebind}. Something that CNN-like transformer architectures \cite{Liu@2021swin,Li@2022MViT,d2021convit,yuan2022volo} cannot do. Finally, while our study primarily focuses on image classification, the implications of ReViT extend beyond this domain. Preliminary experiments on object detection and instance segmentation, using residual attention in spatial-aware multi-scale transformer models such as Swin \cite{Liu@2021swin} and MViTv2 \cite{Li@2022MViT}, hint at its versatility and potential for broader applications in computer vision.

\noindent\textbf{Limitations.} While our study successfully demonstrates a correlation between the globality of attention and feature diversity, it is important to note several limitations. Firstly, residual attention is primarily aimed at mitigating the globalization of the attention mechanism, which has proven beneficial in preserving feature diversity and enhancing overall performance on medium-to-small-scale datasets aligned with our goals. However, its generalization to larger and more diverse networks and datasets remains uncertain. Furthermore, while aiming to preserve the simplicity of self-attention for its ability to generalize across different modalities, the residual attention mechanism we designed was specifically evaluated on image inputs. Although effective for image-based modalities, its generalization to other types of data remains uncertain. 

\noindent\textbf{Future Works.} While this study has offered valuable insights into methods for mitigating feature collapsing, our findings and limitations suggest several promising avenues for future research and development. Firstly, we aim to validate the generalizability of our approach across larger scales, diverse modalities, and more varied tasks to directly address our identified limitations. It would be interesting to apply ReViT on domains such as medical imaging \cite{zhu2021hard} or aerial imaging \cite{kulkarni2023aerial} to observe the effectiveness of ReViT across domains. Secondly, our current residual attention mechanism straightforwardly combats over-globalization by leveraging attention maps derived from diverse feature interactions. In future work, we intend to shift focus towards identifying dominant tokens within sequences that may contribute to feature unification, offering a new perspective on addressing this challenge. Lastly, it would be interesting to investigate how controlling attention globality and feature diversity impacts the explainability of vision transformers.

\section*{Acknowledgements}
This work was supported by "Smart unmannEd AeRial vehiCles for Human likE monitoRing (SEARCHER)" project of the Italian Ministry of Defence within the PNRM 2020 Program (Grant Number: PNRM a2020.231); "A Brain Computer Interface (BCI) based System for Transferring Human Emotions inside Unmanned Aerial Vehicles (UAVs)" Sapienza University Research Projects (Grant Number: RM1221816C1CF63B);
"EYE-FI.AI: going bEYond computEr vision paradigm using wi-FI signals in AI systems" project of the Italian Ministry of Universities and Research (MUR) within the PRIN 2022 Program (Grant Number: 2022AL45R2) (CUP: B53D23012950001); and MICS (Made in Italy – Circular and Sustainable) Extended Partnership and received funding from Next-Generation EU (Italian PNRR – M4 C2, Invest 1.3 – D.D. 1551.11-10-2022, PE00000004). CUP MICS B53C22004130001.
%










\bibliographystyle{cas-model2-names}

\bibliography{cas-refs}

\begin{thebibliography}{47}
\expandafter\ifx\csname natexlab\endcsname\relax\def\natexlab#1{#1}\fi
\providecommand{\url}[1]{\texttt{#1}}
\providecommand{\href}[2]{#2}
\providecommand{\path}[1]{#1}
\providecommand{\DOIprefix}{doi:}
\providecommand{\ArXivprefix}{arXiv:}
\providecommand{\URLprefix}{URL: }
\providecommand{\Pubmedprefix}{pmid:}
\providecommand{\doi}[1]{\href{http://dx.doi.org/#1}{\path{#1}}}
\providecommand{\Pubmed}[1]{\href{pmid:#1}{\path{#1}}}
\providecommand{\bibinfo}[2]{#2}
\ifx\xfnm\relax \def\xfnm[#1]{\unskip,\space#1}\fi
\bibitem[{Cai and Vasconcelos(2018)}]{cascade}
\bibinfo{author}{Cai, Z.}, \bibinfo{author}{Vasconcelos, N.}, \bibinfo{year}{2018}.
\newblock \bibinfo{title}{Cascade r-cnn: Delving into high quality object detection}, in: \bibinfo{booktitle}{IEEE/CVF Conference on Computer Vision and Pattern Recognition (CVPR)}, pp. \bibinfo{pages}{6154--6162}.
\newblock \DOIprefix\doi{10.1109/CVPR.2018.00644}.
\bibitem[{Chen et~al.(2022a)Chen, Zhou, Wang, Dong, Li, Ma, Wang and Zhou}]{chen2022@swipnet}
\bibinfo{author}{Chen, L.}, \bibinfo{author}{Zhou, F.}, \bibinfo{author}{Wang, S.}, \bibinfo{author}{Dong, J.}, \bibinfo{author}{Li, N.}, \bibinfo{author}{Ma, H.}, \bibinfo{author}{Wang, X.}, \bibinfo{author}{Zhou, H.}, \bibinfo{year}{2022}a.
\newblock \bibinfo{title}{Swipenet: Object detection in noisy underwater scenes}.
\newblock \bibinfo{journal}{Pattern Recognition} \bibinfo{volume}{132}, \bibinfo{pages}{108926}.
\newblock \DOIprefix\doi{https://doi.org/10.1016/j.patcog.2022.108926}.
\bibitem[{Chen et~al.(2022b)Chen, Wu, Wang, Hu, Hu, Ding, Cheng and Wang}]{qiang2022mixformer}
\bibinfo{author}{Chen, Q.}, \bibinfo{author}{Wu, Q.}, \bibinfo{author}{Wang, J.}, \bibinfo{author}{Hu, Q.}, \bibinfo{author}{Hu, T.}, \bibinfo{author}{Ding, E.}, \bibinfo{author}{Cheng, J.}, \bibinfo{author}{Wang, J.}, \bibinfo{year}{2022}b.
\newblock \bibinfo{title}{Mixformer: Mixing features across windows and dimensions}, in: \bibinfo{booktitle}{IEEE/CVF Conference on Computer Vision and Pattern Recognition (CVPR)}, pp. \bibinfo{pages}{5239--5249}.
\newblock \DOIprefix\doi{10.1109/CVPR52688.2022.00518}.
\bibitem[{Chu et~al.(2021)Chu, Tian, Wang, Zhang, Ren, Wei, Xia and Shen}]{chu2021twins}
\bibinfo{author}{Chu, X.}, \bibinfo{author}{Tian, Z.}, \bibinfo{author}{Wang, Y.}, \bibinfo{author}{Zhang, B.}, \bibinfo{author}{Ren, H.}, \bibinfo{author}{Wei, X.}, \bibinfo{author}{Xia, H.}, \bibinfo{author}{Shen, C.}, \bibinfo{year}{2021}.
\newblock \bibinfo{title}{Twins: Revisiting the design of spatial attention in vision transformers}, in: \bibinfo{booktitle}{Advances in Neural Information Processing Systems (NeurIPS)}, pp. \bibinfo{pages}{9355--9366}.
\bibitem[{Dai et~al.(2021)Dai, Liu, Le and Tan}]{dai2021coatnet}
\bibinfo{author}{Dai, Z.}, \bibinfo{author}{Liu, H.}, \bibinfo{author}{Le, Q.V.}, \bibinfo{author}{Tan, M.}, \bibinfo{year}{2021}.
\newblock \bibinfo{title}{Coatnet: Marrying convolution and attention for all data sizes}.
\newblock \bibinfo{journal}{Advances in Neural Information Processing Systems (NeurIPS)} \bibinfo{volume}{34}, \bibinfo{pages}{3965--3977}.
\bibitem[{Dosovitskiy et~al.(2021)Dosovitskiy, Beyer, Kolesnikov, Weissenborn, Zhai, Unterthiner, Dehghani, Minderer, Heigold, Gelly, Uszkoreit and Houlsby}]{dosovitskiy2021ViT}
\bibinfo{author}{Dosovitskiy, A.}, \bibinfo{author}{Beyer, L.}, \bibinfo{author}{Kolesnikov, A.}, \bibinfo{author}{Weissenborn, D.}, \bibinfo{author}{Zhai, X.}, \bibinfo{author}{Unterthiner, T.}, \bibinfo{author}{Dehghani, M.}, \bibinfo{author}{Minderer, M.}, \bibinfo{author}{Heigold, G.}, \bibinfo{author}{Gelly, S.}, \bibinfo{author}{Uszkoreit, J.}, \bibinfo{author}{Houlsby, N.}, \bibinfo{year}{2021}.
\newblock \bibinfo{title}{An image is worth 16x16 words: Transformers for image recognition at scale}, in: \bibinfo{booktitle}{International Conference on Learning Representations (ICLR)}, pp. \bibinfo{pages}{1--21}.
\bibitem[{d’Ascoli et~al.(2021)d’Ascoli, Touvron, Leavitt, Morcos, Biroli and Sagun}]{d2021convit}
\bibinfo{author}{d’Ascoli, S.}, \bibinfo{author}{Touvron, H.}, \bibinfo{author}{Leavitt, M.L.}, \bibinfo{author}{Morcos, A.S.}, \bibinfo{author}{Biroli, G.}, \bibinfo{author}{Sagun, L.}, \bibinfo{year}{2021}.
\newblock \bibinfo{title}{Convit: Improving vision transformers with soft convolutional inductive biases}, in: \bibinfo{booktitle}{International Conference on Machine Learning (ICML)}, pp. \bibinfo{pages}{2286--2296}.
\bibitem[{Gao et~al.(2023)Gao, Chen, Liu, Jiang, Li and Ning}]{gao2023@tracking}
\bibinfo{author}{Gao, L.}, \bibinfo{author}{Chen, L.}, \bibinfo{author}{Liu, P.}, \bibinfo{author}{Jiang, Y.}, \bibinfo{author}{Li, Y.}, \bibinfo{author}{Ning, J.}, \bibinfo{year}{2023}.
\newblock \bibinfo{title}{Transformer-based visual object tracking via fine-coarse concatenated attention and cross concatenated mlp}.
\newblock \bibinfo{journal}{Pattern Recognition} , \bibinfo{pages}{109964}\DOIprefix\doi{https://doi.org/10.1016/j.patcog.2023.109964}.
\bibitem[{Girdhar et~al.(2023)Girdhar, El-Nouby, Liu, Singh, Alwala, Joulin and Misra}]{girdhar2023imagebind}
\bibinfo{author}{Girdhar, R.}, \bibinfo{author}{El-Nouby, A.}, \bibinfo{author}{Liu, Z.}, \bibinfo{author}{Singh, M.}, \bibinfo{author}{Alwala, K.V.}, \bibinfo{author}{Joulin, A.}, \bibinfo{author}{Misra, I.}, \bibinfo{year}{2023}.
\newblock \bibinfo{title}{Imagebind: One embedding space to bind them all}, in: \bibinfo{booktitle}{IEEE/CVF Conference on Computer Vision and Pattern Recognition (CVPR)}, pp. \bibinfo{pages}{15180--15190}.
\bibitem[{González-Díaz et~al.(2016)González-Díaz, Buso and Benois-Pineau}]{gonzales2016@visual}
\bibinfo{author}{González-Díaz, I.}, \bibinfo{author}{Buso, V.}, \bibinfo{author}{Benois-Pineau, J.}, \bibinfo{year}{2016}.
\newblock \bibinfo{title}{Perceptual modeling in the problem of active object recognition in visual scenes}.
\newblock \bibinfo{journal}{Pattern Recognition} \bibinfo{volume}{56}, \bibinfo{pages}{129--141}.
\newblock \DOIprefix\doi{https://doi.org/10.1016/j.patcog.2016.03.007}.
\bibitem[{Han et~al.(2021a)Han, Xiao, Wu, Guo, Xu and Wang}]{han2021TnT}
\bibinfo{author}{Han, K.}, \bibinfo{author}{Xiao, A.}, \bibinfo{author}{Wu, E.}, \bibinfo{author}{Guo, J.}, \bibinfo{author}{Xu, C.}, \bibinfo{author}{Wang, Y.}, \bibinfo{year}{2021}a.
\newblock \bibinfo{title}{Transformer in transformer}, in: \bibinfo{booktitle}{Advances in Neural Information Processing Systems (NeurIPS)}, pp. \bibinfo{pages}{15908--15919}.
\bibitem[{Han et~al.(2021b)Han, Xiao, Wu, Guo, Xu and Wang}]{han2021transformer}
\bibinfo{author}{Han, K.}, \bibinfo{author}{Xiao, A.}, \bibinfo{author}{Wu, E.}, \bibinfo{author}{Guo, J.}, \bibinfo{author}{Xu, C.}, \bibinfo{author}{Wang, Y.}, \bibinfo{year}{2021}b.
\newblock \bibinfo{title}{Transformer in transformer}.
\newblock \bibinfo{journal}{Advances in Neural Information Processing Systems (NeurIPS)} \bibinfo{volume}{34}, \bibinfo{pages}{15908--15919}.
\bibitem[{He et~al.(2017)He, Gkioxari, Dollár and Girshick}]{he@2017mask}
\bibinfo{author}{He, K.}, \bibinfo{author}{Gkioxari, G.}, \bibinfo{author}{Dollár, P.}, \bibinfo{author}{Girshick, R.}, \bibinfo{year}{2017}.
\newblock \bibinfo{title}{Mask r-cnn}, in: \bibinfo{booktitle}{IEEE International Conference on Computer Vision (ICCV)}, pp. \bibinfo{pages}{2980--2988}.
\newblock \DOIprefix\doi{10.1109/ICCV.2017.322}.
\bibitem[{He et~al.(2016)He, Zhang, Ren and Sun}]{he2016ResNet}
\bibinfo{author}{He, K.}, \bibinfo{author}{Zhang, X.}, \bibinfo{author}{Ren, S.}, \bibinfo{author}{Sun, J.}, \bibinfo{year}{2016}.
\newblock \bibinfo{title}{Deep residual learning for image recognition}, in: \bibinfo{booktitle}{IEEE/CVF conference on computer vision and pattern recognition (CVPR)}, pp. \bibinfo{pages}{770--778}.
\bibitem[{Hong et~al.(2022)Hong, Pan, Jia, Sun and Gao}]{yuanduo@2022resDnet}
\bibinfo{author}{Hong, Y.}, \bibinfo{author}{Pan, H.}, \bibinfo{author}{Jia, Y.}, \bibinfo{author}{Sun, W.}, \bibinfo{author}{Gao, H.}, \bibinfo{year}{2022}.
\newblock \bibinfo{title}{Resdnet: Efficient dense multi-scale representations with residual learning for high-level vision tasks}.
\newblock \bibinfo{journal}{IEEE Transactions on Neural Networks and Learning Systems (TNLS)} , \bibinfo{pages}{1--12}\DOIprefix\doi{10.1109/TNNLS.2022.3169779}.
\bibitem[{Huff et~al.(2022)Huff, Mahabadi and Tadi}]{huff2021neuroanatomy}
\bibinfo{author}{Huff, T.}, \bibinfo{author}{Mahabadi, N.}, \bibinfo{author}{Tadi, P.}, \bibinfo{year}{2022}.
\newblock \bibinfo{title}{Neuroanatomy, visual cortex}, in: \bibinfo{booktitle}{StatPearls [Internet]}. \bibinfo{publisher}{StatPearls Publishing}, pp.~\bibinfo{pages}{--}.
\bibitem[{Kingma and Ba(2015)}]{kingma@adam}
\bibinfo{author}{Kingma, D.P.}, \bibinfo{author}{Ba, J.}, \bibinfo{year}{2015}.
\newblock \bibinfo{title}{Adam: A method for stochastic optimization.}, in: \bibinfo{booktitle}{International Conference on Learning Representations (ICLR)}, pp. \bibinfo{pages}{1--13}.
\bibitem[{Krizhevsky and Hinton(2009)}]{krizhevsky2009learning}
\bibinfo{author}{Krizhevsky, A.}, \bibinfo{author}{Hinton, G.}, \bibinfo{year}{2009}.
\newblock \bibinfo{title}{Learning multiple layers of features from tiny images}.
\newblock \bibinfo{journal}{Technical Report} .
\bibitem[{Kulkarni and Murala(2023)}]{kulkarni2023aerial}
\bibinfo{author}{Kulkarni, A.}, \bibinfo{author}{Murala, S.}, \bibinfo{year}{2023}.
\newblock \bibinfo{title}{Aerial image dehazing with attentive deformable transformers}, in: \bibinfo{booktitle}{IEEE/CVF Winter Conference on Applications of Computer Vision (WACV)}, pp. \bibinfo{pages}{6305--6314}.
\bibitem[{Laith et~al.(2021)Laith, Jinglan, Amjad~J., Ayad, Ye, Omran, Mohammed~A., Muthana and Laith}]{laith2021cnn}
\bibinfo{author}{Laith, A.}, \bibinfo{author}{Jinglan, Z.}, \bibinfo{author}{Amjad~J., H.}, \bibinfo{author}{Ayad, A.D.}, \bibinfo{author}{Ye, D.}, \bibinfo{author}{Omran, Al-Shamma~J., S.}, \bibinfo{author}{Mohammed~A., F.}, \bibinfo{author}{Muthana, A.A.}, \bibinfo{author}{Laith, F.}, \bibinfo{year}{2021}.
\newblock \bibinfo{title}{Review of deep learning: concepts, cnn architectures, challenges, applications, future directions}.
\newblock \bibinfo{journal}{Journal of Big Data} \bibinfo{volume}{8}, \bibinfo{pages}{1--74}.
\newblock \DOIprefix\doi{10.1186/s40537-021-00444-8}.
\bibitem[{Li et~al.(2022a)Li, Yu, Wang, Yuan, Song and Chen}]{li2022locality}
\bibinfo{author}{Li, K.}, \bibinfo{author}{Yu, R.}, \bibinfo{author}{Wang, Z.}, \bibinfo{author}{Yuan, L.}, \bibinfo{author}{Song, G.}, \bibinfo{author}{Chen, J.}, \bibinfo{year}{2022}a.
\newblock \bibinfo{title}{Locality guidance for improving vision transformers on tiny datasets}, in: \bibinfo{booktitle}{European Conference on Computer Vision (ECCV))}, \bibinfo{organization}{Springer}. pp. \bibinfo{pages}{110--127}.
\bibitem[{Li et~al.(2022b)Li, Wu, Fan, Mangalam, Xiong, Malik and Feichtenhofer}]{Li@2022MViT}
\bibinfo{author}{Li, Y.}, \bibinfo{author}{Wu, C.Y.}, \bibinfo{author}{Fan, H.}, \bibinfo{author}{Mangalam, K.}, \bibinfo{author}{Xiong, B.}, \bibinfo{author}{Malik, J.}, \bibinfo{author}{Feichtenhofer, C.}, \bibinfo{year}{2022}b.
\newblock \bibinfo{title}{Mvitv2: Improved multiscale vision transformers for classification and detection}, in: \bibinfo{booktitle}{IEEE/CVF Conference on Computer Vision and Pattern Recognition (CVPR)}, pp. \bibinfo{pages}{4794--4804}.
\newblock \DOIprefix\doi{10.1109/CVPR52688.2022.00476}.
\bibitem[{Li et~al.(2022c)Li, Liu, Yang, Peng and Zhou}]{li2021survey}
\bibinfo{author}{Li, Z.}, \bibinfo{author}{Liu, F.}, \bibinfo{author}{Yang, W.}, \bibinfo{author}{Peng, S.}, \bibinfo{author}{Zhou, J.}, \bibinfo{year}{2022}c.
\newblock \bibinfo{title}{A survey of convolutional neural networks: Analysis, applications, and prospects}.
\newblock \bibinfo{journal}{IEEE Transactions on Neural Networks and Learning Systems (TNLS)} \bibinfo{volume}{33}, \bibinfo{pages}{6999--7019}.
\newblock \DOIprefix\doi{10.1109/TNNLS.2021.3084827}.
\bibitem[{Lin et~al.(2022)Lin, Cheng, Wu and Shen}]{lin2022cat}
\bibinfo{author}{Lin, H.}, \bibinfo{author}{Cheng, X.}, \bibinfo{author}{Wu, X.}, \bibinfo{author}{Shen, D.}, \bibinfo{year}{2022}.
\newblock \bibinfo{title}{Cat: Cross attention in vision transformer}, in: \bibinfo{booktitle}{IEEE International Conference on Multimedia and Expo (ICME)}, pp. \bibinfo{pages}{1--6}.
\newblock \DOIprefix\doi{10.1109/ICME52920.2022.9859720}.
\bibitem[{Lin et~al.(2014)Lin, Maire, Belongie, Hays, Perona, Ramanan, Dollar and Zitnick}]{lin2014coco}
\bibinfo{author}{Lin, T.Y.}, \bibinfo{author}{Maire, M.}, \bibinfo{author}{Belongie, S.}, \bibinfo{author}{Hays, J.}, \bibinfo{author}{Perona, P.}, \bibinfo{author}{Ramanan, D.}, \bibinfo{author}{Dollar, P.}, \bibinfo{author}{Zitnick, C.L.}, \bibinfo{year}{2014}.
\newblock \bibinfo{title}{Microsoft coco: Common objects in context}, in: \bibinfo{booktitle}{European Conference in Computer Vision (ECCV)}, pp. \bibinfo{pages}{740--755}.
\newblock \DOIprefix\doi{https://doi.org/10.1007/978-3-319-10602-1_48}.
\bibitem[{Liu et~al.(2021)Liu, Lin, Cao, Hu, Wei, Zhang, Lin and Guo}]{Liu@2021swin}
\bibinfo{author}{Liu, Z.}, \bibinfo{author}{Lin, Y.}, \bibinfo{author}{Cao, Y.}, \bibinfo{author}{Hu, H.}, \bibinfo{author}{Wei, Y.}, \bibinfo{author}{Zhang, Z.}, \bibinfo{author}{Lin, S.}, \bibinfo{author}{Guo, B.}, \bibinfo{year}{2021}.
\newblock \bibinfo{title}{Swin transformer: Hierarchical vision transformer using shifted windows}, in: \bibinfo{booktitle}{IEEE/CVF International Conference on Computer Vision (ICCV)}, pp. \bibinfo{pages}{9992--10002}.
\newblock \DOIprefix\doi{10.1109/ICCV48922.2021.00986}.
\bibitem[{Liu et~al.(2022)Liu, Mao, Wu, Feichtenhofer, Darrell and Xie}]{liu2022convnet}
\bibinfo{author}{Liu, Z.}, \bibinfo{author}{Mao, H.}, \bibinfo{author}{Wu, C.Y.}, \bibinfo{author}{Feichtenhofer, C.}, \bibinfo{author}{Darrell, T.}, \bibinfo{author}{Xie, S.}, \bibinfo{year}{2022}.
\newblock \bibinfo{title}{A convnet for the 2020s}, in: \bibinfo{booktitle}{IEEE/CVF Conference on Computer Vision and Pattern Recognition (CVPR)}, pp. \bibinfo{pages}{11976--11986}.
\bibitem[{Loshchilov and Hutter(2019)}]{loshchilov@2018adamW}
\bibinfo{author}{Loshchilov, I.}, \bibinfo{author}{Hutter, F.}, \bibinfo{year}{2019}.
\newblock \bibinfo{title}{Decoupled weight decay regularization}, in: \bibinfo{booktitle}{International Conference on Learning Representations (ICLR)}, pp. \bibinfo{pages}{1--9}.
\bibitem[{Nie et~al.(2024)Nie, Jin, Yan, Chen, Zhu and Qi}]{nie2024scopevit}
\bibinfo{author}{Nie, X.}, \bibinfo{author}{Jin, H.}, \bibinfo{author}{Yan, Y.}, \bibinfo{author}{Chen, X.}, \bibinfo{author}{Zhu, Z.}, \bibinfo{author}{Qi, D.}, \bibinfo{year}{2024}.
\newblock \bibinfo{title}{Scopevit: Scale-aware vision transformer}.
\newblock \bibinfo{journal}{Pattern Recognition} \bibinfo{volume}{153}, \bibinfo{pages}{110470}.
\newblock \DOIprefix\doi{https://doi.org/10.1016/j.patcog.2024.110470}.
\bibitem[{Nilsback and Zisserman(2008)}]{Nilsback2008Flower}
\bibinfo{author}{Nilsback, M.E.}, \bibinfo{author}{Zisserman, A.}, \bibinfo{year}{2008}.
\newblock \bibinfo{title}{Automated flower classification over a large number of classes}, in: \bibinfo{booktitle}{Indian Conference on Computer Vision, Graphics and Image Processing (ICVGIP)}, pp. \bibinfo{pages}{722--729}.
\newblock \DOIprefix\doi{10.1109/ICVGIP.2008.47}.
\bibitem[{Parkhi et~al.(2012)Parkhi, Vedaldi, Zisserman and Jawahar}]{parkhi2012cats}
\bibinfo{author}{Parkhi, O.M.}, \bibinfo{author}{Vedaldi, A.}, \bibinfo{author}{Zisserman, A.}, \bibinfo{author}{Jawahar, C.}, \bibinfo{year}{2012}.
\newblock \bibinfo{title}{Cats and dogs}, in: \bibinfo{booktitle}{IEEE/CVF Conference on Computer Vision and Pattern Recognition (CVPR)}, pp. \bibinfo{pages}{3498--3505}.
\newblock \DOIprefix\doi{10.1109/CVPR.2012.6248092}.
\bibitem[{Russakovsky et~al.(2015)Russakovsky, Deng, Su, Krause, Satheesh, Ma, Huang, Karpathy, Khosla, Bernstein, Berg and Fei-Fei}]{olga2015ImageNet}
\bibinfo{author}{Russakovsky, O.}, \bibinfo{author}{Deng, J.}, \bibinfo{author}{Su, H.}, \bibinfo{author}{Krause, J.}, \bibinfo{author}{Satheesh, S.}, \bibinfo{author}{Ma, S.}, \bibinfo{author}{Huang, Z.}, \bibinfo{author}{Karpathy, A.}, \bibinfo{author}{Khosla, A.}, \bibinfo{author}{Bernstein, M.}, \bibinfo{author}{Berg, A.C.}, \bibinfo{author}{Fei-Fei, L.}, \bibinfo{year}{2015}.
\newblock \bibinfo{title}{Imagenet large scale visual recognition challenge}.
\newblock \bibinfo{journal}{International Journal of Computer Vision (IJCV)} \bibinfo{volume}{115}, \bibinfo{pages}{211--252}.
\newblock \DOIprefix\doi{10.1007/s11263-015-0816-y}.
\bibitem[{Selvaraju et~al.(2017)Selvaraju, Cogswell, Das, Vedantam, Parikh and Batra}]{ramprasaath2017@gcam}
\bibinfo{author}{Selvaraju, R.R.}, \bibinfo{author}{Cogswell, M.}, \bibinfo{author}{Das, A.}, \bibinfo{author}{Vedantam, R.}, \bibinfo{author}{Parikh, D.}, \bibinfo{author}{Batra, D.}, \bibinfo{year}{2017}.
\newblock \bibinfo{title}{Grad-cam: Visual explanations from deep networks via gradient-based localization}, in: \bibinfo{booktitle}{IEEE International Conference on Computer Vision (ICCV)}, pp. \bibinfo{pages}{618--626}.
\newblock \DOIprefix\doi{10.1109/ICCV.2017.74}.
\bibitem[{Tang et~al.(2024)Tang, Lu, Liu, Zhou and Zhang}]{tang2024CATNet}
\bibinfo{author}{Tang, S.}, \bibinfo{author}{Lu, T.}, \bibinfo{author}{Liu, X.}, \bibinfo{author}{Zhou, H.}, \bibinfo{author}{Zhang, Y.}, \bibinfo{year}{2024}.
\newblock \bibinfo{title}{Catnet: Convolutional attention and transformer for monocular depth estimation}.
\newblock \bibinfo{journal}{Pattern Recognition} \bibinfo{volume}{145}, \bibinfo{pages}{109982}.
\newblock \DOIprefix\doi{https://doi.org/10.1016/j.patcog.2023.109982}.
\bibitem[{Tang et~al.(2021)Tang, Han, Xu, Xiao, Deng, Xu and Wang}]{tang2021augmented}
\bibinfo{author}{Tang, Y.}, \bibinfo{author}{Han, K.}, \bibinfo{author}{Xu, C.}, \bibinfo{author}{Xiao, A.}, \bibinfo{author}{Deng, Y.}, \bibinfo{author}{Xu, C.}, \bibinfo{author}{Wang, Y.}, \bibinfo{year}{2021}.
\newblock \bibinfo{title}{Augmented shortcuts for vision transformers}, in: \bibinfo{booktitle}{Advances in Neural Information Processing Systems (NeurIPS)}, pp. \bibinfo{pages}{15316--15327}.
\bibitem[{Touvron et~al.(2021)Touvron, Cord, Douze, Massa, Sablayrolles and Jegou}]{hugo@aug}
\bibinfo{author}{Touvron, H.}, \bibinfo{author}{Cord, M.}, \bibinfo{author}{Douze, M.}, \bibinfo{author}{Massa, F.}, \bibinfo{author}{Sablayrolles, A.}, \bibinfo{author}{Jegou, H.}, \bibinfo{year}{2021}.
\newblock \bibinfo{title}{Training data-efficient image transformers \& distillation through attention}, in: \bibinfo{booktitle}{International Conference on Machine Learning (ICML)}, pp. \bibinfo{pages}{10347--10357}.
\bibitem[{Vaswani et~al.(2017)Vaswani, Shazeer, Parmar, Uszkoreit, Jones, Gomez, Kaiser and Polosukhin}]{ashish@2017attention}
\bibinfo{author}{Vaswani, A.}, \bibinfo{author}{Shazeer, N.}, \bibinfo{author}{Parmar, N.}, \bibinfo{author}{Uszkoreit, J.}, \bibinfo{author}{Jones, L.}, \bibinfo{author}{Gomez, A.N.}, \bibinfo{author}{Kaiser, L.}, \bibinfo{author}{Polosukhin, I.}, \bibinfo{year}{2017}.
\newblock \bibinfo{title}{Attention is all you need}, in: \bibinfo{booktitle}{Advances in Neural Information Processing Systems (NeurIPS)}, pp. \bibinfo{pages}{1--11}.
\bibitem[{Wang et~al.(2017)Wang, Jiang, Qian, Yang, Li, Zhang, Wang and Tang}]{Wang2017res}
\bibinfo{author}{Wang, F.}, \bibinfo{author}{Jiang, M.}, \bibinfo{author}{Qian, C.}, \bibinfo{author}{Yang, S.}, \bibinfo{author}{Li, C.}, \bibinfo{author}{Zhang, H.}, \bibinfo{author}{Wang, X.}, \bibinfo{author}{Tang, X.}, \bibinfo{year}{2017}.
\newblock \bibinfo{title}{Residual attention network for image classification}, in: \bibinfo{booktitle}{IEEE/CVF Conference on Computer Vision and Pattern Recognition (CVPR)}, pp. \bibinfo{pages}{6450--6458}.
\newblock \DOIprefix\doi{10.1109/CVPR.2017.683}.
\bibitem[{Wang et~al.(2022)Wang, Wang, Wang, Lin, Chang, Li and Jin}]{wang2022kvt}
\bibinfo{author}{Wang, P.}, \bibinfo{author}{Wang, X.}, \bibinfo{author}{Wang, F.}, \bibinfo{author}{Lin, M.}, \bibinfo{author}{Chang, S.}, \bibinfo{author}{Li, H.}, \bibinfo{author}{Jin, R.}, \bibinfo{year}{2022}.
\newblock \bibinfo{title}{Kvt: k-nn attention for boosting vision transformers}, in: \bibinfo{booktitle}{European Conference in Computer Vision (ECCV)}, pp. \bibinfo{pages}{285--302}.
\newblock \DOIprefix\doi{https://doi.org/10.1007/978-3-031-19769-7_1}.
\bibitem[{Yao et~al.(2023)Yao, Li, Pan, Wang, Zhang and Mei}]{ting2023dualvit}
\bibinfo{author}{Yao, T.}, \bibinfo{author}{Li, Y.}, \bibinfo{author}{Pan, Y.}, \bibinfo{author}{Wang, Y.}, \bibinfo{author}{Zhang, X.P.}, \bibinfo{author}{Mei, T.}, \bibinfo{year}{2023}.
\newblock \bibinfo{title}{Dual vision transformer}.
\newblock \bibinfo{journal}{IEEE Transactions on Pattern Analysis and Machine Intelligence} \bibinfo{volume}{45}, \bibinfo{pages}{10870--10882}.
\newblock \DOIprefix\doi{10.1109/TPAMI.2023.3268446}.
\bibitem[{Yin et~al.(2021)Yin, Xu, Wang and Zhang}]{yin2021@agunet}
\bibinfo{author}{Yin, Y.}, \bibinfo{author}{Xu, D.}, \bibinfo{author}{Wang, X.}, \bibinfo{author}{Zhang, L.}, \bibinfo{year}{2021}.
\newblock \bibinfo{title}{Agunet: Annotation-guided u-net for fast one-shot video object segmentation}.
\newblock \bibinfo{journal}{Pattern Recognition} \bibinfo{volume}{110}, \bibinfo{pages}{107580}.
\newblock \DOIprefix\doi{https://doi.org/10.1016/j.patcog.2020.107580}.
\bibitem[{Yu et~al.(2022)Yu, Zhao, Li and Yu}]{Yu_2022_BMVC@BOAT}
\bibinfo{author}{Yu, T.}, \bibinfo{author}{Zhao, G.}, \bibinfo{author}{Li, P.}, \bibinfo{author}{Yu, Y.}, \bibinfo{year}{2022}.
\newblock \bibinfo{title}{Boat: Bilateral local attention vision transformer}, in: \bibinfo{booktitle}{British Machine Vision Conference (BMVC)}, pp. \bibinfo{pages}{21--24}.
\bibitem[{Yuan et~al.(2021a)Yuan, Chen, Wang, Yu, Shi, Jiang, Tay, Feng and Yan}]{yuan2021T2T}
\bibinfo{author}{Yuan, L.}, \bibinfo{author}{Chen, Y.}, \bibinfo{author}{Wang, T.}, \bibinfo{author}{Yu, W.}, \bibinfo{author}{Shi, Y.}, \bibinfo{author}{Jiang, Z.}, \bibinfo{author}{Tay, F.E.H.}, \bibinfo{author}{Feng, J.}, \bibinfo{author}{Yan, S.}, \bibinfo{year}{2021}a.
\newblock \bibinfo{title}{Tokens-to-token vit: Training vision transformers from scratch on imagenet}, in: \bibinfo{booktitle}{IEEE/CVF International Conference on Computer Vision (ICCV)}, pp. \bibinfo{pages}{538--547}.
\newblock \DOIprefix\doi{10.1109/ICCV48922.2021.00060}.
\bibitem[{Yuan et~al.(2021b)Yuan, Chen, Wang, Yu, Shi, Jiang, Tay, Feng and Yan}]{yuan2021tokens}
\bibinfo{author}{Yuan, L.}, \bibinfo{author}{Chen, Y.}, \bibinfo{author}{Wang, T.}, \bibinfo{author}{Yu, W.}, \bibinfo{author}{Shi, Y.}, \bibinfo{author}{Jiang, Z.H.}, \bibinfo{author}{Tay, F.E.}, \bibinfo{author}{Feng, J.}, \bibinfo{author}{Yan, S.}, \bibinfo{year}{2021}b.
\newblock \bibinfo{title}{Tokens-to-token vit: Training vision transformers from scratch on imagenet}, in: \bibinfo{booktitle}{Proceedings of the IEEE/CVF international conference on computer vision}, pp. \bibinfo{pages}{558--567}.
\bibitem[{Yuan et~al.(2022)Yuan, Hou, Jiang, Feng and Yan}]{yuan2022volo}
\bibinfo{author}{Yuan, L.}, \bibinfo{author}{Hou, Q.}, \bibinfo{author}{Jiang, Z.}, \bibinfo{author}{Feng, J.}, \bibinfo{author}{Yan, S.}, \bibinfo{year}{2022}.
\newblock \bibinfo{title}{Volo: Vision outlooker for visual recognition}.
\newblock \bibinfo{journal}{IEEE Transactions on Pattern Analysis and Machine Intelligence} \bibinfo{volume}{45}, \bibinfo{pages}{6575--6586}.
\bibitem[{Zheng et~al.(2017)Zheng, Zhao, Gao and Wu}]{zheng2017@classification}
\bibinfo{author}{Zheng, P.}, \bibinfo{author}{Zhao, Z.Q.}, \bibinfo{author}{Gao, J.}, \bibinfo{author}{Wu, X.}, \bibinfo{year}{2017}.
\newblock \bibinfo{title}{Image set classification based on cooperative sparse representation}.
\newblock \bibinfo{journal}{Pattern Recognition} \bibinfo{volume}{63}, \bibinfo{pages}{206--217}.
\newblock \DOIprefix\doi{https://doi.org/10.1016/j.patcog.2016.09.043}.
\bibitem[{Zhu et~al.(2021)Zhu, Chen, Peng, Wang and Jin}]{zhu2021hard}
\bibinfo{author}{Zhu, C.}, \bibinfo{author}{Chen, W.}, \bibinfo{author}{Peng, T.}, \bibinfo{author}{Wang, Y.}, \bibinfo{author}{Jin, M.}, \bibinfo{year}{2021}.
\newblock \bibinfo{title}{Hard sample aware noise robust learning for histopathology image classification}.
\newblock \bibinfo{journal}{IEEE Transactions on Medical Imaging} \bibinfo{volume}{41}, \bibinfo{pages}{881--894}.

\end{thebibliography}



\end{document}